\documentclass[10pt,twocolumn,letterpaper]{article}

\usepackage{iccv}              

\definecolor{iccvblue}{rgb}{0.21,0.49,0.74}
\usepackage[pagebackref,breaklinks,colorlinks,allcolors=iccvblue]{hyperref}
\usepackage{multirow}
\usepackage{color}
\usepackage{colortbl}

\def\paperID{4} 
\def\confName{ICCV}
\def\confYear{2025}

\title{On the Use of Hierarchical Vision Foundation Models\\for Low-Cost Human Mesh Recovery and Pose Estimation}

\author{Shuhei Tarashima$^{\dag\ddag}$\\
{\tt\small tarashima@acm.org}
\and
Yushan Wang$^{\ddag}$\\
{\tt\small yushanwang218@gmail.com}
\and
Norio Tagawa$^{\ddag}$\\
{\tt\small tagawa@tmu.ac.jp}\\
\and
$^{\dag}$ NTT DOCOMO Business
\and
$^{\ddag}$ Tokyo Metropolitan University
}

\begin{document}
\twocolumn[{%
\renewcommand\twocolumn[1][]{#1}%
\maketitle
\includegraphics[width=.5\linewidth]{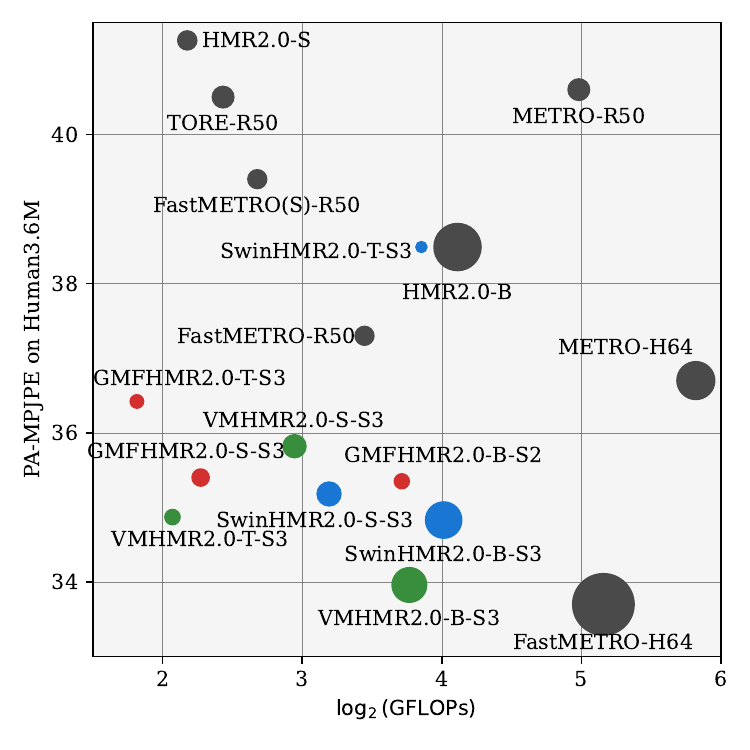}
\includegraphics[width=.5\linewidth]{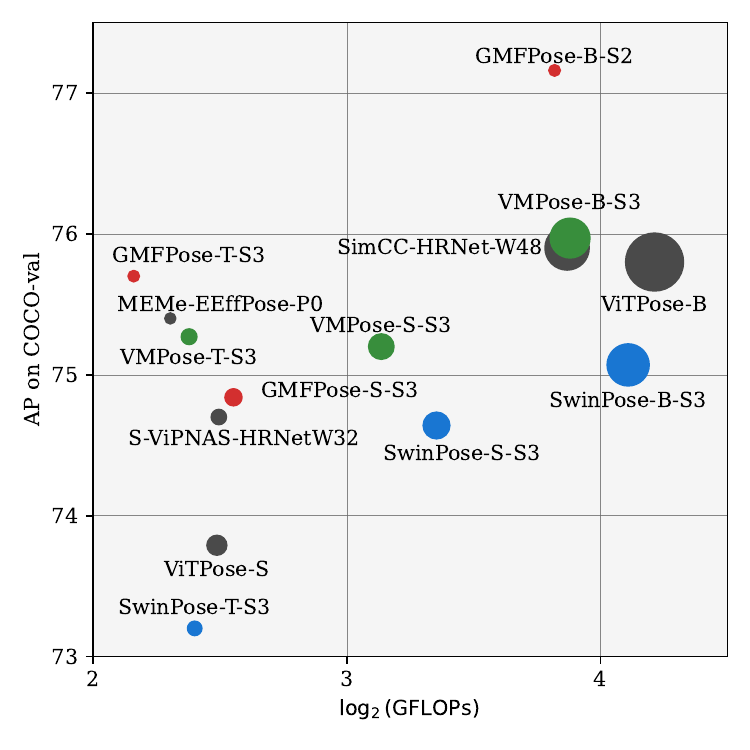}
\vspace{-2em}
\captionof{figure}{
(Left): 3D pose estimation results (PA-MPJPE; lower is better) of human mesh recovery (HMR) models on the Human3.6M dataset \cite{ionescu+2014tpami}.
(Right): 2D pose estimation results (AP; higher is better) of human pose estimation (HPE) models on the COCO-val dataset \cite{lin+2014eccv}.
Circle size reflects model size. 
Gray circles indicate existing methods, while colored circles represent our proposed models, which leverage the early stages of hierarchical vision foundation models (VFMs) as encoders.
These results demonstrate that our models tend to offer a more favorable trade-off between performance and computational cost ({\it i.e.}, GFLOPs) compared to existing approaches.
}
\vspace{1.5em}
\label{fig:teaser}
}]
\begin{abstract}
In this work, we aim to develop simple and efficient models for human mesh recovery (HMR) and its predecessor task, human pose estimation (HPE).
State-of-the-art HMR methods, such as HMR2.0 and its successors, rely on large, non-hierarchical vision transformers as encoders, which are inherited from the corresponding HPE models like ViTPose.
To establish baselines across varying computational budgets, we first construct three lightweight HMR2.0 variants by adapting the corresponding ViTPose models.
In addition, we propose leveraging the early stages of hierarchical vision foundation models (VFMs), including Swin Transformer, GroupMixFormer, and VMamba, as encoders.
This design is motivated by the observation that intermediate stages of hierarchical VFMs produce feature maps with resolutions comparable to or higher than those of non-hierarchical counterparts.
We conduct a comprehensive evaluation of 27 hierarchical-VFM-based HMR and HPE models, demonstrating that using only the first two or three stages achieves performance on par with full-stage models.
Moreover, we show that the resulting truncated models exhibit better trade-offs between accuracy and computational efficiency compared to existing lightweight alternatives.
The source code is available at \url{https://github.com/nttcom/TruncHierVFM}.
\end{abstract}    
\section{Introduction}
\label{sec:intro}
Human mesh recovery (HMR) plays a central role in a wide range of applications, including animation, virtual try-on, sports analytics, and human-computer interaction \cite{zheng+2023acm,tian+2023tpami,gao+2025air,liu+2024neuro}.
Over the past decade, this research field has witnessed remarkable progress, driven in part by vision foundation models (VFMs) \cite{he+2016cvpr,wang+2021tpami,sun+2019cvpr,dosovitskiy+2020iclr}.
While early HMR approaches \cite{kanazawa+2018cvpr,zhang+2021iccv,kolotouros+2021iccv,li+2022eccv2,kocabas+2021iccv} primarily relied on convolutional neural networks (CNNs), recent state-of-the-art (SoTA) methods \cite{cho+2022eccv,dou+2023iccv,lin+2021cvpr,lin+2021iccv,agarwal+2024eccvw,pang+2022neurips,cai+2024tpami,romain+2025cvprw,lin+2025cvpr,feng+2024cvpr} have increasingly adopted Transformer-based architectures \cite{vaswani+2017nips}.
Among them, HMR2.0 \cite{goel+2023iccv} has garnered significant attention for its simplicity and strong performance.
HMR2.0 and its successors \cite{patel+20243dv,fiche+2025cvpr,saleem+2025aaai,prospero+2025cvprw,stathopoulos+2024cvpr,dwivedi+2024cvpr} employ a large non-hierarchical vision transformer ({\it i.e.}, ViT-H \cite{dosovitskiy+2020iclr}) as their encoder, which is inherited from the corresponding human pose estimation (HPE) model ({\it i.e.}, ViTPose-H \cite{xu+2022neurips}).
\par
In general, even for HMR and HPE, large VFMs demand substantial computational resources, which can hinder their deployment in real-time or resource-constrained settings such as mobile devices or edge computing environments.
With HMR2.0, a straightforward approach to alleviate this issue is to use smaller ViT variants ({\it e.g.}, ViT-L, ViT-B, ViT-S) as encoders.
Since this direction has not been explored in the literature, in this work we instantiate smaller variants of HMR2.0 as baselines (see \S\ref{sec:baseline:hmr2:small} for details).
Beyond this, to better balance performance and efficiency while preserving the architectural simplicity of HMR2.0, we investigate the use of {\it hierarchical} VFMs \cite{liu+2021iccv,ge+2023arxiv,liu+2024neurips} as encoders for HMR and its predecessor, HPE.
The key insight motivating our approach lies in the resolution characteristics of intermediate representations in hierarchical VFMs.
They typically follow a four-stage structure, where the spatial resolution of feature maps are higher or the same with the consistent resolution seen in non-hierarchical VFMs.
Therefore, if the intermediate outputs of pretrained hierarchical VFMs retain sufficient semantic richness and spatial detail, the latter stages of the original backbone can be removed.
This allows for reductions in model size and computational cost without compromising architectural simplicity.
\par
We conduct extensive experiments to validate this observation by instantiating HMR and HPE models within the HMR2.0 and ViTPose frameworks, using different stages of three hierarchical VFMs as encoders: Swin Transformer \cite{liu+2021iccv}, GroupMixFormer \cite{ge+2024cvpr}, and VMamba \cite{liu+2024neurips}.
In total, we instantiate 27 hierarchical-VFM-based HMR and HPE models. 
Our results consistently show that models using only the first two or three VFM stages as encoders achieve performance comparable to, and occasionally better than, their full four-stage counterparts.
Moreover, these {\it truncated} models demonstrate a more favorable trade-off between accuracy and efficiency compared to existing lightweight approaches, including the small ViT-based variants of HMR2.0 and ViTPose.
Our contributions are summarized as follows:
\begin{itemize}
\item We instantiate lightweight variants of HMR2.0 via inheriting the encoders of ViTPose-\{L,B,S\}.
\item We investigate the use of hierarchical VFMs \cite{liu+2021iccv,ge+2023arxiv,liu+2024neurips} as encoders within the HMR2.0 and ViTPose frameworks. Experimental results show that models utilizing only the first few VFM stages as their encoders achieve performance comparable to, and sometimes better than, their full four-stage counterparts for both HMR and HPE.
\item We further demonstrate that hierarchical-VFM-based HMR and HPE models achieve superior trade-offs between performance and efficiency compared to existing lightweight approaches, including those based on ViT.
\end{itemize}
\begin{figure}[t]
\centering
\begin{minipage}[b]{1.0\linewidth}
    \centering
\includegraphics[keepaspectratio, width=\linewidth, page=3]{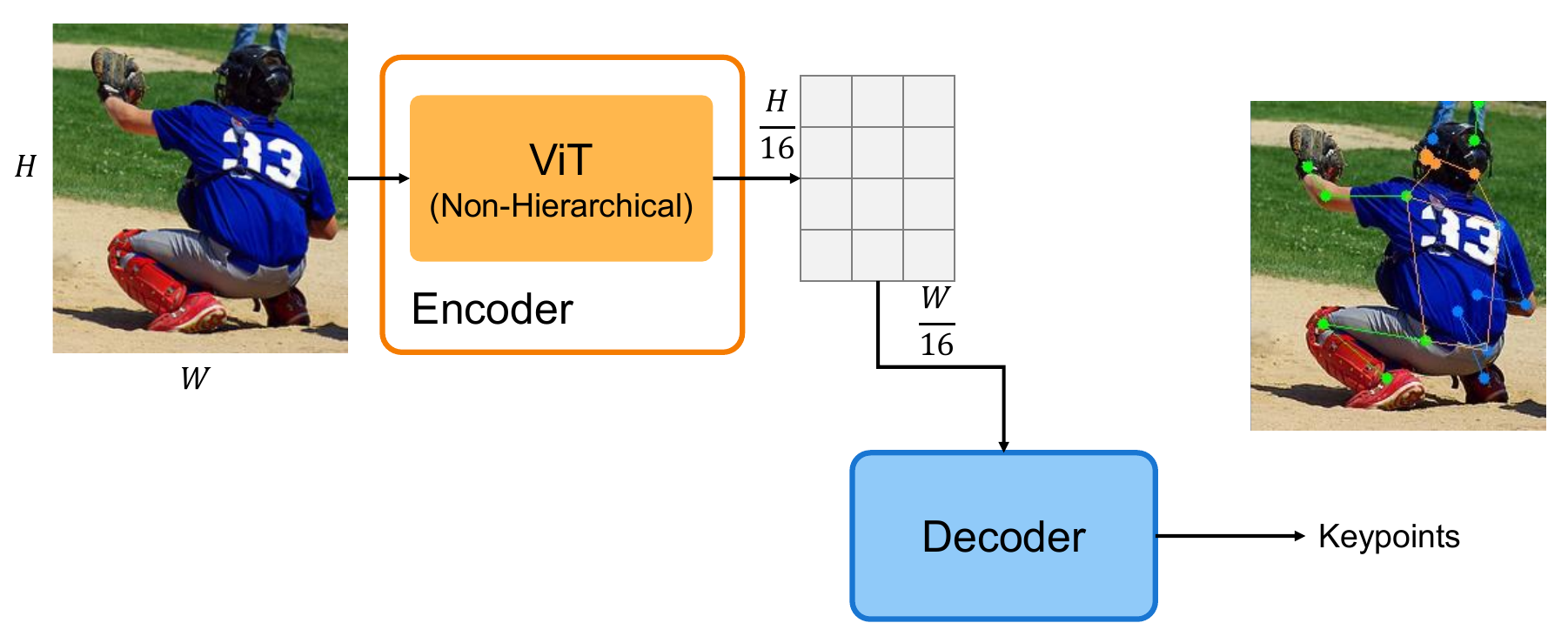}
    \caption*{(a) ViTPose \cite{xu+2022neurips}}
  \end{minipage} \\
  \begin{minipage}[b]{1.0\linewidth}
    \centering
\includegraphics[keepaspectratio, width=\linewidth, page=2]{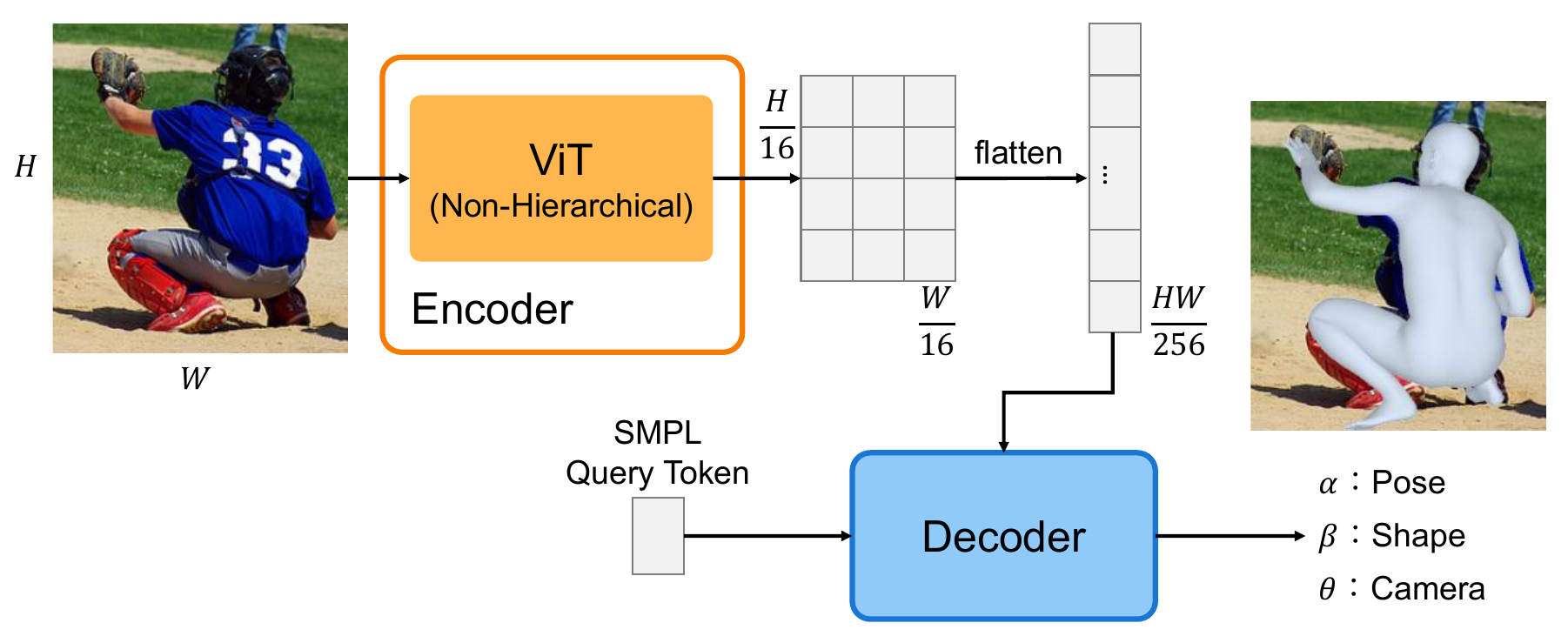}
    \caption*{(b) HMR2.0 \cite{goel+2023iccv}}
  \end{minipage}
\caption{
Architectures of the baseline models for human pose estimation (HPE) and human mesh recovery (HMR).
}
\label{fig:baseline}
\end{figure}

\section{Related Work}
\label{sec:related}
\subsection{Human Mesh Recovery (HMR)}
\label{sec:related:hmr}
HMR has been extensively studied under a variety of problem settings, including multi-person HMR \cite{romain+2025cvprw, liao+2024tvcg, sun+2021iccv, sun+2022cvpr, su+2025cvpr, fabien+2024eccv, sun+2024cvpr, choi+2022cvpr, zanfir+2018cvpr, zanfir+2018neurips, qiu+2023cvpr}, multi-view HMR \cite{bogo+eccv2016, zhang+2023iccv, zhu+2025aaai, li+2021wacv, pavlakos+2022cvpr, yu+2022neurips, jia+2023aaai, xie+2024icmr}, video-based HMR \cite{zhang+2025cvpr, choi+2021cvpr, kocabas+2020cvpr, wei+2022cvpr, kang+2025cvprw, shen+2024sa, yang+2023iclr}, full-body HMR \cite{pavlakos+2019cvpr, wang+2022cvpr, shen+2024acmmm, zhang+2023tpami, li+2025tpami,lin+2023cvpr,cai+2023neurips}, privacy-preserving HMR \cite{gao+2025air, wu+2025cvpr, ge+2024cvpr}, and prompt-based HMR \cite{wang+2025cvpr, lin+2025cvpr, feng+2024cvpr}.
In this work, we focus on the most fundamental setting, image-based HMR, where the objective is to predict SMPL \cite{loper+2015tog} parameters from a single cropped image of a person.
\par
While early works employed CNNs such as ResNet \cite{he+2016cvpr}, HRNet \cite{sun+2019cvpr,wang+2021tpami}, and EfficientNet \cite{tan+2019icml} as backbone encoders \cite{fiche+2024eccv,le+2024cvpr,ma+2023cvpr,kanazawa+2018cvpr,zhang+2021iccv,kolotouros+2021iccv,shen+2025icml,sarandi+2024neurips,xiao+2024eccv,li+2022eccv2,kocabas+2021iccv,li+2024tip,xu+2024cvpr}, they have been superseded by Transformer-based methods in recent years. 
For instance, METRO \cite{lin+2021cvpr,cho+2022eccv}, Mesh Graphormer \cite{lin+2021iccv} and PointHMR \cite{kim+2023cvpr} introduce hybrid CNN-Transformer architectures that leverage pretrained CNN features while capturing global dependencies via self-attention. 
On the other hand, several recent works \cite{goel+2023iccv,lin+2023cvpr,cai+2024tpami,cai+2023neurips} utilize fully Transformer-based encoders.
HMR2.0 \cite{goel+2023iccv}, along with its successors \cite{patel+20243dv,fiche+2025cvpr,saleem+2025aaai,prospero+2025cvprw,stathopoulos+2024cvpr,dwivedi+2024cvpr}, employs ViT-H as the encoder, initialized through pretraining on HPE tasks.
SMPLer-X \cite{cai+2023neurips}, uses four ViTs to construct full-body HMR models with varying model sizes.
More recently, prompt-based HMR approaches including ChatPose \cite{feng+2024cvpr} and ChatHuman \cite{lin+2025cvpr} leverage the ViT encoder from CLIP \cite{radford+2021icml} to align visual and textual features.
\par
Previous works have primarily explored the use of non-hierarchical transformers, ViT or DeiT, for the HMR task. 
In contrast, this study investigates the application of hierarchical VFMs, encompassing not only transformer-based architectures but also recently proposed state space models (SSMs).
A notable exception in the literature is \cite{pang+2022neurips}, which integrates hierarchical transformers such as the Swin Transformer \cite{liu+2021iccv, liu+2022cvpr} and Twins \cite{chu+2021neurips} into the HMR framework of \cite{kanazawa+2018cvpr}.
DeFormer \cite{yoshiyasu+2023cvpr} also explores the use of hierarchical transformers, such as the Swin Transformer with a Feature Pyramid Network and the Mix Transformer \cite{cao+2022ijcai,xie+2021neurips}, as backbones.
Nevertheless, as discussed in \S\ref{sec:intro}, our work goes beyond simply applying hierarchical VFMs to HMR:
We further investigate the use of only the initial stages of these hierarchical models as encoders, aiming to develop more efficient HMR architectures.
\par
Several studies have aimed to develop efficient HMR models \cite{zheng+2023cvpr,dou+2023iccv,agarwal+2024eccvw,wang+2025visapp}. 
For instance, CoarseMETRO \cite{agarwal+2024eccvw} employs a coarse-to-fine strategy to reduce the computational burden of early transformer layers, while TORE \cite{dou+2023iccv} accelerates HMR by pruning background tokens.
POTTER \cite{zheng+2023cvpr} integrates a high-resolution stream with a basic stream to recover more accurate human meshes while reducing memory usage and computational cost.
Although effective, these methods often introduce additional architectural complexity. 
In contrast, our approach, leveraging the early stages of hierarchical VFMs, offers a simpler alternative that is potentially complementary to these techniques.
\subsection{Human Pose Estimation (HPE)}
\label{sec:related:hpe}
Since the encoders in HMR methods are often inherited from those used in HPE models, another promising direction for developing efficient HMR is to adapt efficient HPE approaches.
Several studies have explored this in the context of HPE. 
For example, MEMe \cite{li+2022sensors}, Lite-HRNet \cite{yu+2021cvpr}, and LitePose \cite{wang+2022cvpr} focus on optimizing popular CNN backbones such as EfficientNet \cite{tan+2019icml} and HRNet \cite{sun+2019cvpr, wang+2021tpami} to improve the trade-off between accuracy and efficiency.
DANet \cite{luo+2021icme} proposes an improved multi-scale feature fusion strategy that eliminates the need for computationally expensive cascaded pyramid architectures. 
SimCC \cite{li+2022eccv} reformulates HPE as two independent classification tasks for horizontal and vertical coordinates.
Additionally, CNF \cite{yang+2022pr} and ViPNAS \cite{xu+2021cvpr} apply neural architecture search to automatically optimize network structures for improved efficiency.
\par
All of the aforementioned approaches are based on CNN architectures. 
While CNNs can be considered a type of hierarchical VFM, in this work we focus on adapting more recent architectures, such as transformers and SSMs.
\subsection{Vision Foundation Model (VFM)}
\label{sec:related:vfm}
VFMs can be broadly categorized into non-hierarchical models \cite{dosovitskiy+2020iclr,touvron+2021icml,steiner+2022tmlr,beyer+2023cvpr,fang+2024ivc,tian+2023cvpr,wei+2023iccv,yang+2023iclr2,alkin+2025iclr,yang+2022pr} and hierarchical models \cite{liu+2021iccv,fan+2021iccv,wang+2021iccv,zhang+2021iccv,li+2022cvpr,wang+2022cvmj,sun+2023tpami,zhang+2023iclr,ge+2023arxiv,liu+2024neurips}.
Following the discussion above, this work focuses on recent hierarchical VFMs.
Among them, we select Swin Transformer (Swin) \cite{liu+2021iccv}, GroupMixFormer (GMF), and VMamba (VM) \cite{liu+2022cvpr} based on their strong performance, widespread recognition, and the availability of public code with pretrained weights.
Please refer to the original papers for detailed architectural descriptions.
All three models consist of four VFM stages, with output feature map resolutions of $1/4 \times 1/4$, $1/8 \times 1/8$, $1/16 \times 1/16$, and $1/32 \times 1/32$ relative to the input image resolution at stages 1 through 4, respectively.
In \S\ref{sec:proposed}, we will explore to use their first few stages as HMR and HPE encoders.
\begin{table}[t]
\centering
\scalebox{0.76}{
\begin{tabular}{ll|cccc}
\toprule
                  &  & HMR2.0 \cite{goel+2023iccv} & HMR2.0-L & HMR2.0-B & HMR2.0-S \\ \hline
\multirow{2}{*}{Encoder} & & ViTPose-H & ViTPose-L & ViTPose-B & ViTPose-S \\
                  &  & (631.0 M) & (303.3 M) & (85.8 M) & (21.7 M) \\ \hline
\multirow{5}{*}{Decoder} & $N$ & 6 & 6 & 3 & 3 \\
                  & $h$ & 8 & 8 & 8 & 4 \\
                  & $d_{\text{hid}}$ & 64 & 32 & 24 & 16 \\
                  & $d_{\text{ff}}$ & 1024 & 512 & 384 & 128 \\
                  &  & (39.5 M) & (19.1 M) & (7.0 M) & (2.3 M) \\ \hline
total             &  & (670.5M) & (322.4M) & (92.8M) & (24.0M) \\ 
\bottomrule
\end{tabular}
}
\caption{
Building on the original HMR2.0 architecture proposed by \cite{goel+2023iccv}, we introduce three scaled variants: HMR2.0-L, HMR2.0-B, and HMR2.0-S.
$N$ denotes the number of Transformer layers, $h$ represents the number of attention heads in each cross-attention layer, $d_{\text{hid}}$ is the hidden dimension size, and $d_{\text{ff}}$ refers to the hidden dimension size of the feed-forward MLP block.
}
\label{tab:hmr:variants}
\end{table}
\begin{figure*}[t]
\centering
\begin{minipage}[b]{0.32\linewidth}
    \centering
\includegraphics[keepaspectratio, width=\linewidth, page=3]{figures/enc_hier.pdf}
    \caption*{(a)}
  \end{minipage} 
  \begin{minipage}[b]{0.32\linewidth}
    \centering
\includegraphics[keepaspectratio, width=\linewidth, page=2]{figures/enc_hier.pdf}
    \caption*{(b)}
  \end{minipage} 
    \begin{minipage}[b]{0.32\linewidth}
    \centering
\includegraphics[keepaspectratio, width=\linewidth, page=1]{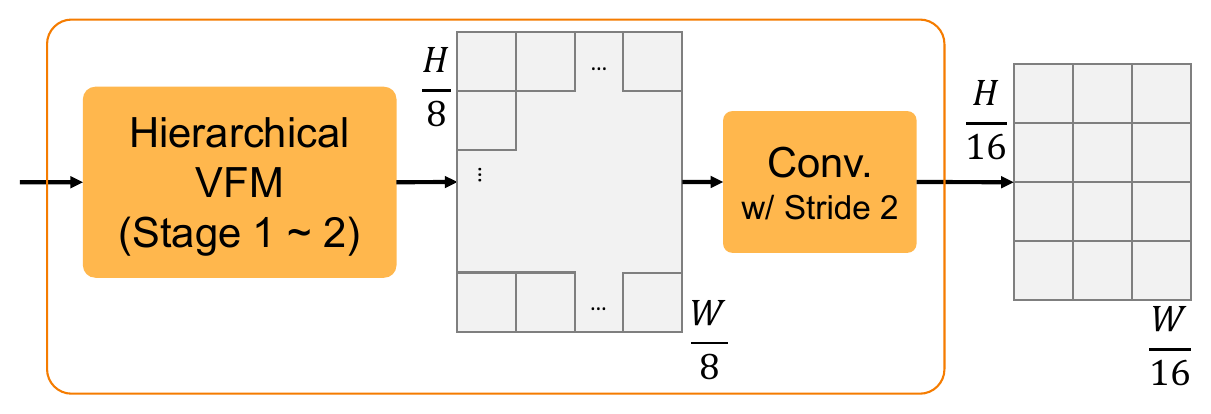}
    \caption*{(c)}
  \end{minipage}
\caption{
We explore the efficient utilization of hierarchical VFMs as encoders for HMR and HPE.
(a) When using all four stages, the output feature resolution of stage 4 is $1/2 \times 1/2$ relative to that of non-hierarchical VFMs. To match the expected resolution, we apply a deconvolution layer with stride 2 to upsample the features.
(b) When using up to stage 3, the output resolution matches that of non-hierarchical VFMs, so we directly feed the features into the decoder without additional processing.
(c) When using up to stage 2, we apply a convolution layer with stride 2 to downsample the features to the target resolution.
}
\label{fig:hier}
\end{figure*}

\begin{table*}[htbp]
\begin{minipage}[c]{\hsize}
\centering
\scalebox{0.95}{
\begin{tabular}{l|rrr|rr|rr|rr|rr|rr|rr}
\toprule
 & \multicolumn{3}{c|}{Up to Stage 4 (S4)} & \multicolumn{6}{c|}{Up to Stage 3 (S3)} & \multicolumn{6}{c}{Up to Stage 2 (S2)} \\
 & P & F & $\Phi^{\text{P,2D}}$ & P & $\Delta$ & F & $\Delta$ & $\Phi^{\text{P,2D}}$ & $\Delta$ & P & $\Delta$ & F & $\Delta$ & $\Phi^{\text{P,2D}}$ & $\Delta$ \\ \midrule
SwinPose-B & 92.0 & 19.1 & 77.8 & \cellcolor{green!15}\color{red}62.6 & \cellcolor{green!15}-32.0 & \cellcolor{green!15}\color{red}17.3 & \cellcolor{green!15}-9.7 & \cellcolor{green!15}\color{blue}77.7 & \cellcolor{green!15}-0.1 & \color{red}5.8 & -93.7 & \color{red}4.2 & -78.2 & \color{blue}61.2 & -21.4 \\
SwinPose-S & 52.6 & 11.3 & 77.3 & \cellcolor{green!15}\color{red}36.1 & \cellcolor{green!15}-31.4 & \cellcolor{green!15}\color{red}10.2 & \cellcolor{green!15}-9.3 & \cellcolor{green!15}\color{red}77.4 & \cellcolor{green!15}0.1 & \color{red}4.1 & -92.2 & \color{red}2.8 & -75.1 & \color{blue}56.8 & -26.5 \\
SwinPose-T & 31.3 & 6.3 & 76.3 & \cellcolor{green!15}\color{red}14.8 & \cellcolor{green!15}-52.8 & \cellcolor{green!15}\color{red}5.3 & \cellcolor{green!15}-16.6 & \cellcolor{green!15}\color{blue}75.9 & \cellcolor{green!15}-0.6 & \color{red}4.1 & -86.8 & \color{red}2.8 & -55.6 & \color{blue}57.7 & -24.4 \\
GMFPose-B & 48.3 & 18.3 & 79.7 & \color{red}24.9 & -48.5 & \color{red}17.1 & -6.4 & \color{blue}79.6 & -0.2 & \cellcolor{green!15}\color{red}9.9 & \cellcolor{green!15}-79.6 & \cellcolor{green!15}\color{red}14.1 & \cellcolor{green!15}-22.9 & \cellcolor{green!15}\color{blue}79.5 & \cellcolor{green!15}-0.2 \\
GMFPose-S & 24.9 & 6.2 & 77.7 & \cellcolor{green!15}\color{red}19.4 & \cellcolor{green!15}-22.2 & \cellcolor{green!15}\color{red}5.9 & \cellcolor{green!15}-4.6 & \cellcolor{green!15}\color{blue}77.4 & \cellcolor{green!15}-0.4 & \color{red}4.3 & -82.9 & \color{red}2.8 & -53.8 & \color{blue}74.3 & -4.4 \\
GMFPose-T & 12.8 & 4.6 & 78.3 & \cellcolor{green!15}\color{red}9.7 & \cellcolor{green!15}-24.1 & \cellcolor{green!15}\color{red}4.5 & \cellcolor{green!15}-3.5 & \cellcolor{green!15}78.3 & \cellcolor{green!15}0.0 & \color{red}3.8 & -70.5 & \color{red}3.3 & -29.2 & \color{blue}75.3 & -3.8 \\
VMPose-B & 92.8 & 16.4 & 78.5 & \cellcolor{green!15}\color{red}58.6 & \cellcolor{green!15}-36.9 & \cellcolor{green!15}\color{red}14.7 & \cellcolor{green!15}-10.1 & \cellcolor{green!15}\color{red}78.5 & \cellcolor{green!15}0.1 & \color{red}6.2 & -93.3 & \color{red}4.6 & -72.0 & \color{blue}74.7 & -4.8 \\
VMPose-S & 53.2 & 9.7 &	77.8 & \cellcolor{green!15}\color{red}33.9 & \cellcolor{green!15}-36.3 & \cellcolor{green!15}\color{red}8.8 & \cellcolor{green!15}-9.6 & \cellcolor{green!15}77.8 & \cellcolor{green!15}0.0 & \color{red}4.4 & -91.8 & \color{red}3.1 & -68.6 & \color{blue}72.3 & -7.1 \\
VMPose-T & 33.3 & 6.0 & 77.7 & \cellcolor{green!15}\color{red}17.0 & \cellcolor{green!15}-48.9 & \cellcolor{green!15}\color{red}5.2 & \cellcolor{green!15}-13.0 & \cellcolor{green!15}\color{red}77.8 & \cellcolor{green!15}0.1 & \color{red}4.1 & -87.6 & \color{red}2.7 & -54.7 & \color{blue}70.6 & -9.1 \\
\bottomrule
\end{tabular}
}
\caption*{
(a) HPE models.
$\Phi^{\text{P,2D}}$ is the HPE performance score, as defined in Equation \eqref{eq:hpe}.
A higher value of $\Phi^{\text{P,2D}}$ indicates better performance.
}
\label{tab:stages:hpe}
\end{minipage} \\
\begin{minipage}[c]{\hsize}
\scalebox{0.9}{
\begin{tabular}{l|rrr|rr|rr|rr|rr|rr|rr}
\toprule
 & \multicolumn{3}{c|}{S4} & \multicolumn{6}{c|}{S3} & \multicolumn{6}{c}{S2} \\
 & P & F & $\Phi^{\text{M,2D}}$ & P & $\Delta$ & F & $\Delta$ & $\Phi^{\text{M,2D}}$ & $\Delta$ & P & $\Delta$ & F & $\Delta$ & $\Phi^{\text{M,2D}}$ & $\Delta$ \\ \midrule
SwinHMR2.0-B & 95.5 & 18.0 & 80.2 & \cellcolor{green!15}\color{red}66.1 & \cellcolor{green!15}-30.8 & \cellcolor{green!15}\color{red}16.2 & \cellcolor{green!15}-10.3 & \cellcolor{green!15}\color{blue}79.9 & \cellcolor{green!15}-0.5 & \color{red}9.3 & -90.2 & \color{red}3.1 & -82.9 & \color{blue}66.1 & -17.7\\
SwinHMR2.0-S & 52.3 & 10.2 & 80.0 & \cellcolor{green!15}\color{red}35.8 & \cellcolor{green!15}-31.6 & \cellcolor{green!15}\color{red}9.1 & \cellcolor{green!15}-10.2 & \cellcolor{green!15}\color{blue}78.9 & \cellcolor{green!15}-1.4 & \color{red}3.8 & -92.7 & \color{red}1.7 & -83.0 & \color{blue}60.4 & -24.5 \\
SwinHMR2.0-T & 31.0 & 5.2 & 77.7 & \cellcolor{green!15}\color{red}14.5 & \cellcolor{green!15}-53.4 & \cellcolor{green!15}\color{red}4.2 & \cellcolor{green!15}-19.8 & \cellcolor{green!15}\color{blue}75.7 & \cellcolor{green!15}-2.5 & \color{red}3.8 & -87.7 & \color{red}1.7 & -67.0 & \color{blue}61.3 & -21.1 \\
GMFHMR2.0-B & 52.4 & 17.3 & 82.1 & \color{red}29.0 & -44.7 & \color{red}16.1 & -6.8 & \color{blue}81.5 & -0.7 & \cellcolor{green!15}\color{red}14.0 & \cellcolor{green!15}-73.4 & \cellcolor{green!15}\color{red}13.1 & \cellcolor{green!15}-24.2 & \cellcolor{green!15}\color{blue}79.3 & \cellcolor{green!15}-3.4\\
GMFHMR2.0-S & 24.8 & 5.1 & 80.6 & \cellcolor{green!15}\color{red}19.3 & \cellcolor{green!15}-22.2 & \cellcolor{green!15}\color{red}4.8 & \cellcolor{green!15}-5.5 & \cellcolor{green!15}\color{blue}79.9 & \cellcolor{green!15}-0.8 & \color{red}4.2 & -83.2 & \color{red}1.8 & -64.8 & \color{blue}72.4 & -10.2 \\
GMFHMR2.0-T & 13.1 & 3.7 & 80.1 & \cellcolor{green!15}\color{red}10.1 & \cellcolor{green!15}-23.5 & \cellcolor{green!15}\color{red}3.5 & \cellcolor{green!15}-4.3 & \cellcolor{green!15}\color{blue}79.0 & \cellcolor{green!15}-1.4 & \color{red}4.1 & -68.5 & \color{red}2.3 & -36.7 & \color{blue}72.9 & -9.0 \\
VMHMR2.0-B & 96.3 & 15.3 & 81.2 & \cellcolor{green!15}\color{red}62.1 & \cellcolor{green!15}-35.5 & \cellcolor{green!15}\color{red}13.6 & \cellcolor{green!15}-10.8 & \cellcolor{green!15}\color{blue}80.3 & \cellcolor{green!15}-1.1 & \color{red}9.8 & -89.9 & \color{red}3.5 & -77.2 & \color{blue}72.9 & -10.3 \\
VMHMR2.0-S & 52.9 & 8.6 & 81.0 & \cellcolor{green!15}\color{red}33.6 & \cellcolor{green!15}-36.5 & \cellcolor{green!15}\color{red}7.7 & \cellcolor{green!15}-10.8 & \cellcolor{green!15}\color{blue}80.1 & \cellcolor{green!15}-1.1 & \color{red}4.1 & -92.3 & \color{red}2.0 & -77.3 & \color{blue}68.8 & -15.1 \\
VMHMR2.0-T & 33.0 & 4.9 & 79.6 & \cellcolor{green!15}\color{red}14.5 & \cellcolor{green!15}-56.1 & \cellcolor{green!15}\color{red}4.2 & \cellcolor{green!15}-14.3 & \cellcolor{green!15}\color{blue}78.6 & \cellcolor{green!15}-1.4 & \color{red}3.8 & -88.5 & \color{red}1.6 & -66.9 & \color{blue}67.5 & -15.3 \\
\bottomrule
\end{tabular}
}
\caption*{
(b) HMR models for 2D pose estimation.
$\Phi^{\text{M,2D}}$ indicates the HMR performance score for 2D pose estimation, as defined in Equation \eqref{eq:hmr:2d}.
A higher value of $\Phi^{\text{M,2D}}$ indicates better performance.
}
\label{tab:stages:hmr:2d}
\end{minipage} \\
\begin{minipage}[c]{\hsize}
\scalebox{0.9}{
\begin{tabular}{l|rrr|rr|rr|rr|rr|rr|rr}
\toprule
 & \multicolumn{3}{c|}{S4} & \multicolumn{6}{c|}{S3} & \multicolumn{6}{c}{S2} \\
 & P & F & $\Phi^{\text{M,3D}}$ & P & $\Delta$ & F & $\Delta$ & $\Phi^{\text{M,3D}}$ & $\Delta$ & P & $\Delta$ & F & $\Delta$ & $\Phi^{\text{M,3D}}$ & $\Delta$ \\ \midrule 
SwinHMR2.0-B & 95.5 & 18.0 & 55.6 & \cellcolor{green!15}\color{red}66.1 & \cellcolor{green!15}-30.8 & \cellcolor{green!15}\color{red}16.2 & \cellcolor{green!15}-10.3 & \cellcolor{green!15}\color{red}55.5 & \cellcolor{green!15}-0.2 & \color{red}9.3 & -90.2 & \color{red}3.1 & -82.9 & \color{blue}67.7 & 21.6 \\
SwinHMR2.0-S & 52.3 & 10.2 & 55.7 & \cellcolor{green!15}\color{red}35.8 & \cellcolor{green!15}-31.6 & \cellcolor{green!15}\color{red}9.1 & \cellcolor{green!15}-10.2 & \cellcolor{green!15}\color{red}55.5 & \cellcolor{green!15}-0.3 & \color{red}3.8 & -92.7 & \color{red}1.7 & -83.0 & \color{blue}73.8 & 32.7 \\
SwinHMR2.0-T & 31.0 & 5.2 & 56.8 & \cellcolor{green!15}\color{red}14.5 & \cellcolor{green!15}-53.4 & \cellcolor{green!15}\color{red}4.2 & \cellcolor{green!15}-19.8 & \cellcolor{green!15}\color{blue}57.9 & \cellcolor{green!15}2.0 & \color{red}3.8 & -87.7 & \color{red}1.7 & -67.0 & \color{blue}72.6 & 27.8 \\
GMFHMR2.0-B & 52.4 & 17.3 & 55.0 & \color{red}29.0 & -44.7 & \color{red}16.1 & -6.8 & 55.0 & 0.0 & \cellcolor{green!15}\color{red}14.0 & \cellcolor{green!15}-73.4 & \cellcolor{green!15}\color{red}13.1 & \cellcolor{green!15}-24.2 & \cellcolor{green!15}\color{blue}55.4 & \cellcolor{green!15}0.8 \\
GMFHMR2.0-S & 24.8 & 5.1 & 56.0 & \cellcolor{green!15}\color{red}19.3 & \cellcolor{green!15}-22.2 & \cellcolor{green!15}\color{red}4.8 & \cellcolor{green!15}-5.5 & \cellcolor{green!15}56.0 & \cellcolor{green!15}0.0 & \color{red}4.2 & -83.2 & \color{red}1.8 & -64.8 & \color{blue}60.7 & 7.9 \\
GMFHMR2.0-T & 13.1 & 3.7 & 56.1 & \cellcolor{green!15}\color{red}10.1 & \cellcolor{green!15}-23.5 & \cellcolor{green!15}\color{red}3.5 & \cellcolor{green!15}-4.3 & \cellcolor{green!15}56.1 & \cellcolor{green!15}0.0 & \color{red}4.1 & -68.5 & \color{red}2.3 & -36.7 & \color{blue}59.7 & 6.5 \\
VMHMR2.0-B & 96.3 & 15.3 & 54.6 & \cellcolor{green!15}\color{red}62.1 & \cellcolor{green!15}-35.5 & \cellcolor{green!15}\color{red}13.6 & \cellcolor{green!15}-10.8 & \cellcolor{green!15}\color{blue}55.0 & \cellcolor{green!15}0.7 & \color{red}9.8 & -89.9 & \color{red}3.5 & -77.2 & \color{blue}58.6 & 7.4 \\
VMHMR2.0-S & 52.9 & 8.6 & 55.9 & \cellcolor{green!15}\color{red}33.6 & \cellcolor{green!15}-36.5 & \cellcolor{green!15}\color{red}7.7 & \cellcolor{green!15}-10.8 & \cellcolor{green!15}\color{blue}56.1 & \cellcolor{green!15}0.3 & \color{red}4.1 & -92.3 & \color{red}2.0 & -77.3 & \color{blue}62.3 & 11.4 \\
VMHMR2.0-T & 33.0 & 4.9 & 55.7 & \cellcolor{green!15}\color{red}14.5 & \cellcolor{green!15}-56.1 & \cellcolor{green!15}\color{red}4.2 & \cellcolor{green!15}-14.3 & \cellcolor{green!15}\color{red}55.5 & \cellcolor{green!15}-0.4 & \color{red}3.8 & -88.5 & \color{red}1.6 & -66.9 & \color{blue}64.1 & 15.0 \\
\bottomrule
\end{tabular}
}
\caption*{
(c) HMR models for 3D pose estimation.
$\Phi^{\text{M,3D}}$ indicates the HMR performance score for 3D pose estimation, as defined in Equation \eqref{eq:hmr:3d}.
A lower value of $\Phi^{\text{M,3D}}$ indicates better performance.
}
\label{tab:stages:hmr:3d}
\end{minipage}
\caption{
Performance comparison of HPE/HMR Models with varying encoder stage depths. 
In each table, P denotes the number of model parameters (in millions), and F indicates the computational cost in GFLOPs.
$\Delta$ represents the relative performance change (in percentage) compared to the corresponding full-stage model ({\it i.e.}, S4).
Red text indicates improvement, while blue text denotes degradation compared to the full-stage model.
Models highlighted with green cells are used in the following comparisons with existing methods.
\vspace{-1em}
}
\label{tab:stages}
\end{table*}
\section{Baseline for HMR \& HPE}
\label{sec:baseline}
To ensure better self-containment, we briefly review ViTPose \cite{xu+2022neurips} and HMR2.0 \cite{goel+2023iccv} in this section, as they serve as baselines for HPE and HMR, respectively.
Both ViTPose and HMR2.0 employ a non-hierarchical VFM, ViT \cite{dosovitskiy+2020iclr} as their encoder.
While ViTPose explores four ViT variants of different sizes, HMR2.0 utilizes only the largest variant.
To establish a more comprehensive baseline, we introduce smaller ViT-based variants of HMR2.0 in \S~\ref{sec:baseline:hmr2:small}, using encoders inherited from the corresponding ViTPose models.
\subsection{ViTPose \cite{xu+2022neurips}}
\label{sec:baseline:vitpose}
As illustrated in Figure~\ref{fig:baseline}~(a), ViTPose adopts a straightforward encoder-decoder architecture: a ViT serves as the encoder to generate a feature map, while the decoder comprises deconvolution layers followed by a prediction layer that outputs heatmaps corresponding to keypoints.
Given an image of a person with height $H$ and width $W$ (typically, $H=256$ and $W=192$), the encoder produces features with a spatial resolution of $H/16 \times W/16$.
In this work, we utilize the official ViTPose repository\footnote{\url{https://github.com/ViTAE-Transformer/ViTPose}}, including pretrained weights, to deploy four model variants, {\it i.e.}, ViTPose-H, ViTPose-L, ViTPose-B, and ViTPose-S, all of which are finetuned on the COCO dataset \cite{lin+2014eccv}.
The results, including model size, computational complexity (GFLOPs), and frames per second (FPS), are summarized in Table~\ref{tab:hpe:results:full}.
\subsection{HMR2.0 \cite{goel+2023iccv}}
\label{sec:baseline:hmr2}
HMR2.0 also adopts an encoder-decoder architecture.
As illustrated in Figure~\ref{fig:baseline}~(b), a ViT-based encoder produces a feature map, which is subsequently flattened and passed to a transformer-based decoder to predict the SMPL \cite{loper+2015tog} parameters, {\it i.e.}, the pose parameters $\alpha$, shape parameters $\beta$, and camera parameters $\theta$.
During training, the encoder is initialized with pretrained weights from the ViTPose.
Note that in \cite{goel+2023iccv}, only ViTPose-H is used as the encoder.
Its architectural specifications are provided in Table~\ref{tab:hmr:variants}, and the corresponding 2D and 3D pose estimation results obtained using the official code\footnote{\url{https://github.com/shubham-goel/4D-Humans}} are reported in Table~\ref{tab:hmr:results:full}.
\subsubsection{HMR2.0 with Smaller ViTs}
\label{sec:baseline:hmr2:small}
Smaller ViTPose-based encoders can be integrated into the HMR2.0 framework in a straightforward manner.
Based on the encoder and decoder parameter sizes of the original HMR2.0 with the ViTPose-H encoder, we construct three additional HMR2.0 variants {\it i.e.}, HMR2.0-L, HMR2.0-B and HMR2.0-S.
Their architectural specifications are summarized in Table~\ref{tab:hmr:variants}.
Each model is initialized with pretrained weights from the corresponding ViTPose encoder and trained under the same settings as in \cite{goel+2023iccv}.
\par
The resulting 2D and 3D pose estimation performances are summarized in Table~\ref{tab:hpe:results:full} and \ref{tab:hmr:results:full}.
As expected, the accuracies gradually decrease as the model size becomes smaller.
\section{Hierarchical VFM as HMR / HPE Encoder}
\label{sec:proposed}
In this work, we explore the potential of hierarchical VFMs for efficient HMR and HPE.
To this end, our design is guided by two principles:
(1) maintaining a simple architecture by avoiding complex or highly specialized modules, and
(2) preserving architectural consistency with the corresponding HMR2.0 and ViTPose baselines described in \S\ref{sec:baseline}.
\par
As noted in \S\ref{sec:related:vfm}, most hierarchical VFMs comprise four stages, with the output resolutions of stages 2, 3, and 4 being $2 \times 2$, $1 \times 1$, and $1/2 \times 1/2$ relative to the output resolution of non-hierarchical VFMs, respectively.
Therefore, when utilizing all four stages of a hierarchical VFM, we add a $2 \times 2$ deconvolution layer to align the output resolution with that of non-hierarchical VFMs.
Conversely, when using only the first two stages, we insert a convolutional layer with stride 2 after stage 2 to downsample the feature map.
When using the first three stages, the output from stage 3 is directly fed into the decoder without additional processing.
These configurations are illustrated in Figure~\ref{fig:hier}.
Notice that, to keep our modifications minimal, the number of channels in the added convolutional or deconvolutional layers is matched to the output channel size of stage 3 for each VFM.
\par
Following this principle, we instantiate HMR and HPE models based on the HMR2.0 and ViTPose frameworks using three hierarchical VFMs: Swin \cite{liu+2021iccv}, GMF \cite{ge+2023arxiv}, and VM \cite{liu+2024neurips}.
For clarity, we adopt a consistent naming convention where each model name comprises the encoder identifier, followed by the task identifier ({\it i.e.}, Pose for HPE or HMR2.0 for HMR), the encoder size ({\it i.e.}, Base (B), Small (S), or Tiny (T)), and a suffix indicating the number of encoder stages used.
For example, SwinPose-S-S3 denotes an HPE model that uses the small variant of Swin Transformer up to Stage 3, while VMHMR2.0-T-S2 refers to an HMR model based on the tiny variant of VMamba using up to Stage 2.
For the decoder, we adopt the same architecture as ViTPose or HMR2.0, matched to the corresponding encoder size.
Therefore, models such as GMFPose-B-S4, GMFPose-B-S3, and GMFPose-B-S2 share the same decoder architecture, regardless of the number of encoder stages utilized.
Note that for HPE, we follow the decoder design of the ViTPose variants \cite{xu+2022neurips}.
For HMR, the decoder architecture corresponding to each encoder size ({\it i.e.}, base and small variants of each VFM) is consistent with those defined in Section~\ref{sec:baseline:hmr2:small} (cf. Table~\ref{tab:hmr:variants}).
Additionally, for tiny encoder models, we use the same decoder architecture as that of the small variants.

\begin{figure}[t]
\centering
\includegraphics[keepaspectratio, width=\linewidth, page=2]{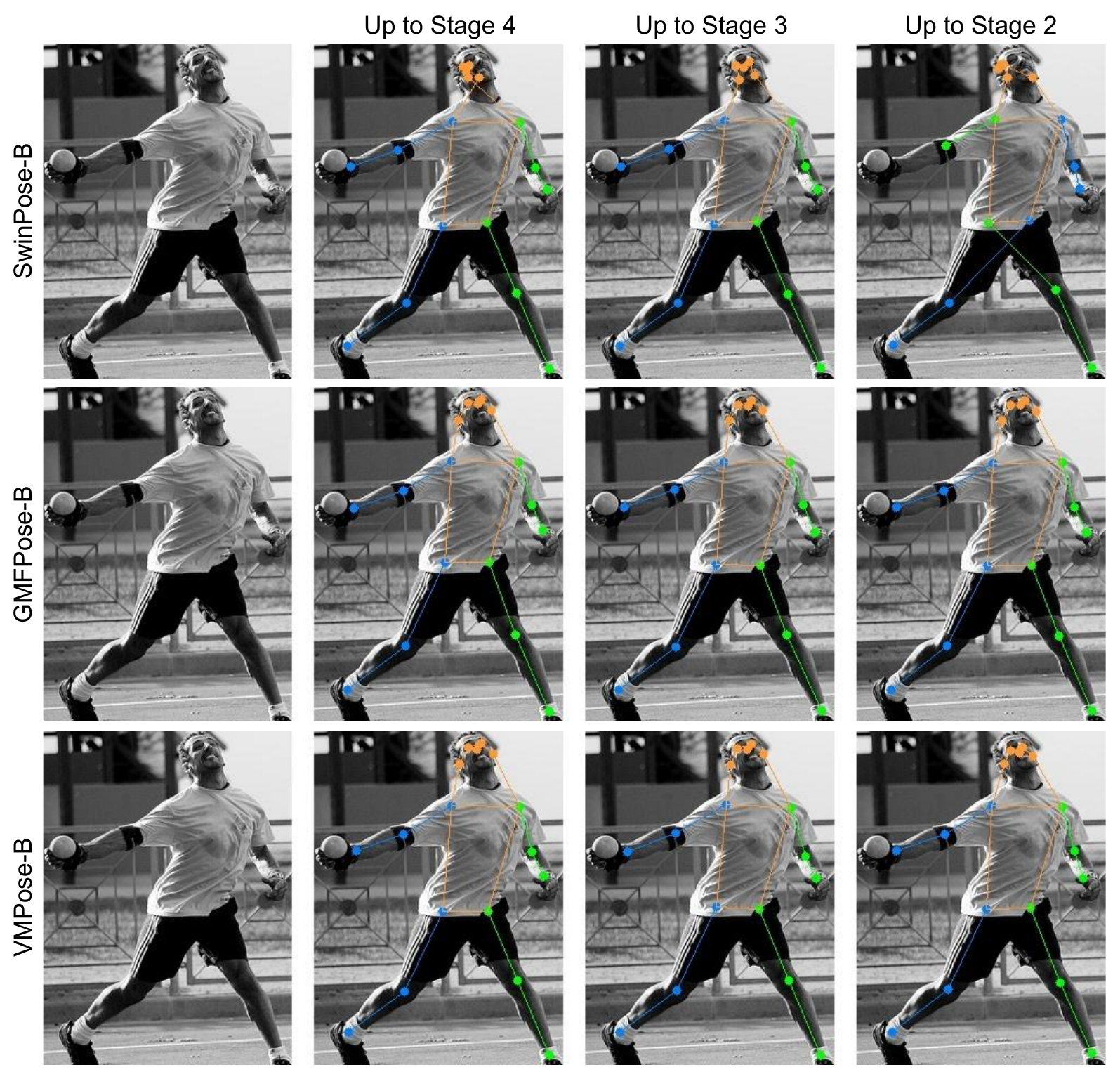}
\caption{
Qualitative results of hierarchical-VFM-based HPE models.
}
\label{fig:qualitative:hpe}
\end{figure}
\section{Evaluation}
\label{sec:eval}
\subsection{Setting}
\label{sec:eval:imple}
We follow the training and evaluation protocols established by our baselines \cite{xu+2022neurips, goel+2023iccv} for HMR and HPE.
For HPE, we use the COCO dataset \cite{lin+2014eccv} for both training and evaluation.
All the hierarchical-VFM-based encoders are initialized with ImageNet-1K pre-trained weights provided by their respective official repositories\footnote{\url{https://github.com/microsoft/Swin-Transformer}}\footnote{\url{https://github.com/AILab-CVC/GroupMixFormer}}\footnote{\url{https://github.com/MzeroMiko/VMamba}}.
HPE performance is evaluated using Average Precision (AP) and Average Recall (AR) metrics.
For HMR, we train our models on a mixed dataset comprising Human3.6M \cite{ionescu+2014tpami}, MPI-INF-3DHP \cite{mehta+20173dv}, COCO \cite{lin+2014eccv}, MPII \cite{andriluka+2014cvpr}, InstaVariety \cite{kanazawa+2019cvpr}, AVA \cite{gu+2018cvpr}, and AI Challenger \cite{wu+2019icme}.
We evaluate 2D pose estimation accuracy using LSP-Extended \cite{johnson+2011cvpr}, COCO-val \cite{lin+2014eccv}, and PoseTrack-val \cite{andriluka+2018cvpr}, reporting the Percentage of Correct Keypoints (PCK) of reprojected keypoints at thresholds 0.05 and 0.1.
For 3D pose accuracy, we use the 3DPW-test \cite{kolotouros+2019iccv} and Human3.6M-val \cite{ionescu+2014tpami} datasets, reporting both Mean Per Joint Position Error (MPJPE) and Procrustes Aligned MPJPE (PA-MPJPE).
All models are trained using 8 A100 GPUs. 
Inference speed, measured in frames per second (FPS), is evaluated using a single A100 GPU.
\par
The results of all hierarchical-VFM-based HPE and HMR models (27 models in total), along with ViT-based models, are presented in Table~\ref{tab:hpe:results:full} and Table~\ref{tab:hmr:results:full}.
Based on these results, the following subsections provide a validation of our key observations and a comparison with existing lightweight models.
\begin{figure}[t]
\centering
\includegraphics[keepaspectratio, width=\linewidth, page=2]{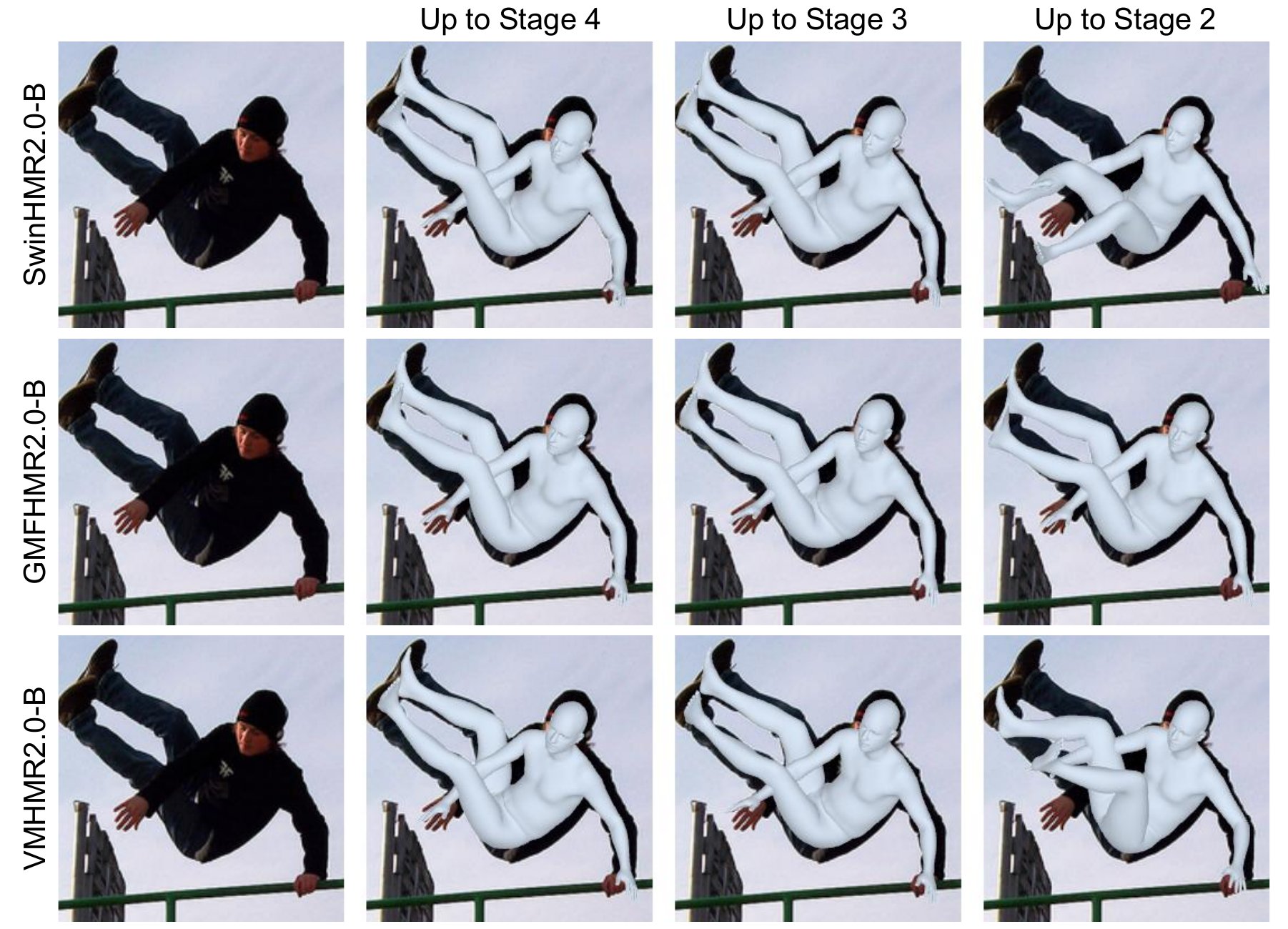}
\caption{
Qualitative results of hierarchical-VFM-based HMR models.
}
\label{fig:qualitative:hmr}
\end{figure}
\subsection{Truncated Hierarchical VFMs}
\label{sec:eval:init}
Here we evaluate our key idea: leveraging only the initial few stages of hierarchical VFMs ({\it i.e.}, truncated hierarchical VFMs) for efficient HPE and HMR.
To enable a simplified and unified comparison, we define three aggregated performance scores: $\Phi^{\text{P,2D}}$ for 2D pose estimation by HPE, $\Phi^{\text{M,2D}}$ for 2D pose estimation by HMR, and $\Phi^{\text{M,3D}}$ for 3D pose estimation by HMR.
Each metric is computed as the average of the respective evaluation scores across relevant datasets.
Formally, the metrics are defined as follows:
\begin{align}
\Phi^{\text{P,2D}} &= \frac{1}{|\mathcal{D}^{\text{P,2D}}|} \sum_{\mathcal{D}^{\text{P,2D}}} \frac{1}{2}(\text{AP} + \text{AR}), \label{eq:hpe} \\
\Phi^{\text{M,2D}} &= \frac{1}{|\mathcal{D}^{\text{M,2D}}|} \sum_{\mathcal{D}^{\text{M,2D}}} \frac{1}{2}(\text{PCK@0.05} + \text{PCK@0.1}), \label{eq:hmr:2d} \\
\Phi^{\text{M,3D}} &= \frac{1}{|\mathcal{D}^{\text{M,3D}}|} \sum_{\mathcal{D}^{\text{M,3D}}} \frac{1}{2}(\text{MPJPE} + \text{PA-MPJPE}), \label{eq:hmr:3d}
\end{align}
where $\mathcal{D}^{\text{P,2D}}$, $\mathcal{D}^{\text{M,2D}}$, and $\mathcal{D}^{\text{M,3D}}$ denote the datasets for evaluating pose estimation by HPE, 2D pose estimation by HMR, and 3D pose estimation by HMR, respectively\footnote{
Specifically, in this study $\mathcal{D}^{\text{HPE,2D}}$ consists of COCO-val \cite{lin+2014eccv},
$\mathcal{D}^{\text{HMR,2D}}$ consists of LSP-Extended \cite{johnson+2011cvpr}, COCO-val \cite{lin+2014eccv}, and PoseTrack-val \cite{andriluka+2018cvpr}, and 
$\mathcal{D}^{\text{HMR,3D}}$ consists of 3DPW-test \cite{kolotouros+2019iccv} and Human3.6M-val \cite{ionescu+2014tpami}.
}.
Note that higher values of $\Phi^{\text{HPE}}$ and $\Phi^{\text{HMR,2D}}$ indicate better performance, while for $\Phi^{\text{HMR,3D}}$, lower values are preferable.
\par
Tables~\ref{tab:stages} (a)–(c) present the model size, computational complexity, and the performance scores defined above for nine hierarchical-VFM-based models, each utilizing a different number of encoder stages.
As expected, using fewer VFM stages leads to reduced model size and lower computational cost for both HPE and HMR models.
Notably, models using up to stage 3 perform comparably to those employing all four stages.
Surprisingly, in some cases such as 2D pose estimation by HPE and 3D pose estimation by HMR, models with only three stages even slightly outperform their full-stage counterparts.
In the case of GMF-B, using just the first two stages still yields performance close to that of the full four-stage model.
\par
These trends are also reflected in the qualitative results shown in Figure~\ref{fig:qualitative:hpe} and \ref{fig:qualitative:hmr}.
Overall, the results confirm that the first two or three stages of hierarchical VFMs serve as efficient and effective encoders for both HPE and HMR.
Based on these findings, we select the models highlighted in green in Table\ref{tab:stages} for comparison with existing methods.
\subsection{Comparison to Existing Methods}
\label{sec:eval:comp:existing}
The left panel of Figure~\ref{fig:teaser} illustrates the 3D pose estimation accuracy of HMR models on the Human3.6M dataset \cite{ionescu+2014tpami}, including selected hierarchical-VFM-based models, ViT-based HMR2.0 variants, and existing approaches \cite{lin+2021cvpr,cho+2022eccv,dou+2023iccv}.
The right panel presents the 2D pose estimation performance of HPE models, namely, hierarchical-VFM-based models, small ViTPose variants, and existing methods \cite{li+2022eccv,li+2022sensors,xu+2021cvpr}, on the COCO dataset \cite{lin+2014eccv}.
In the left panel, models positioned closer to the bottom-left corner are both more accurate and computationally efficient, while in the right panel, those closer to the top-left corner are preferred.
Results in these figures demonstrate that our hierarchical-VFM-based models generally achieve lower PA-MPJPE with smaller model sizes and reduced computational cost for HMR, and offer competitive accuracy for HPE.
Despite their simplicity, the findings suggest that hierarchical-VFM-based models can serve as strong and efficient baselines for both HPE and HMR tasks.

\section{Conclusion}
\label{sec:conc}
In this work, we explore the use of hierarchical VFMs as encoders for HMR and HPE models.
Experimental results show that utilizing only the initial few stages of these models yields HMR/HPE performance that is comparable to, or in some cases slightly better than, that of full-stage counterparts, while successfully reducing model size and computational cost.
Furthermore, these truncated models demonstrate a favorable trade-off between accuracy and efficiency compared to existing lightweight alternatives.
\par
For future work, we plan to extend our study by instantiating additional models based on other hierarchical VFMs.
We also aim to validate our approach on broader HMR tasks, including full-body and multi-person HMR.
\begin{table}[]
\centering
\scalebox{0.87}{
\begin{tabular}{l|rrr|cc}
\toprule
 & & & & \multicolumn{2}{c}{COCO \cite{lin+2014eccv}} \\
 & Param. & GFLOPs & FPS & AP & AR \\ \midrule
\rowcolor{gray!15}
ViTPose-H \cite{xu+2022neurips} & 637.2 & 125.9 & 67 & 79.1 & 84.1 \\
\rowcolor{gray!15}
ViTPose-L \cite{xu+2022neurips} & 308.5 & 61.6 & 126 & 78.3 & 83.5 \\
\rowcolor{gray!15}
ViTPose-B \cite{xu+2022neurips} & 90.0 & 18.5 & 393 & 75.8 & 81.1 \\
\rowcolor{gray!15}
ViTPose-S \cite{xu+2022neurips} & 24.3 & 5.6 & 936 & 73.8 & 79.2 \\ \midrule
SwinPose-B-S4 & 92.0 & 19.1 & 284 & 75.1 & 80.4 \\
SwinPose-B-S3 & 62.6 & 17.3 & 307 & 75.1 & 80.3 \\
SwinPose-B-S2 & 5.8 & 4.2 & 978 & 57.8 & 64.5 \\
SwinPose-S-S4 & 52.6 & 11.3 & 406 & 74.5 & 80.1 \\
SwinPose-S-S3 & 36.1 & 10.2 & 436 & 74.6 & 80.1 \\
SwinPose-S-S2 & 4.1 & 2.8 & 1197 & 53.2 & 60.5 \\
SwinPose-T-S4 & 31.3 & 6.3 & 665 & 73.7 & 79.0 \\
SwinPose-T-S3 & 14.8 & 5.3 & 757 & 73.2 & 78.6 \\
SwinPose-T-S2 & 4.1 & 2.8 & 1185 & 54.2 & 61.3 \\ \midrule
GMFPose-B-S4 & 48.3 & 18.3 & 171 & 77.2 & 82.2 \\
GMFPose-B-S3 & 24.9 & 17.1 & 182 & 77.1	& 82.0 \\
GMFPose-B-S2 & 9.9	& 14.1 & 207 & 77.2 & 81.9 \\
GMFPose-S-S4 & 24.9 & 6.2 & 480 & 75.2 & 80.3 \\
GMFPose-S-S3 & 19.4 & 5.9 & 511 & 74.8 & 80.0 \\
GMFPose-S-S2 & 4.3 & 2.8 & 869 & 71.8 & 76.9 \\
GMFPose-T-S4 & 12.8 & 4.6 & 445 & 75.8 & 80.9 \\
GMFPose-T-S3 & 9.7 & 4.5 & 468 & 75.7 & 80.9 \\
GMFPose-T-S2 & 3.8 & 3.3 & 619 & 72.8 & 77.9 \\ \midrule
VMPose-B-S4 & 92.8 & 16.4 & 469 & 75.9 & 81.0 \\
VMPose-B-S3 & 58.6 & 14.7 & 532 & 76.0 & 81.1 \\
VMPose-B-S2 & 6.2 & 4.6 & 1046 & 72.1 & 77.3 \\
VMPose-S-S4 & 53.2 & 9.7 & 585 & 75.2 & 80.4 \\
VMPose-S-S3 & 33.9 & 8.8 & 660 & 75.2 & 80.3 \\
VMPose-S-S2 & 4.4 & 3.1 & 1234 & 69.6 & 74.9 \\
VMPose-T-S4 & 33.3 & 6.0 & 951 & 75.2 & 80.3 \\
VMPose-T-S3 & 17.0 & 5.2 & 1067 & 75.3 & 80.4 \\
VMPose-T-S2 & 4.1 & 2.7 & 1518 & 67.9 & 73.4 \\
\bottomrule
\end{tabular}
}
\caption{
Pose estimation results of HPE models on the COCO-val dataset \cite{lin+2014eccv}.
Models shaded with gray cells are evaluated using publicly available pretrained weights, while the others are trained in our environment.
}
\label{tab:hpe:results:full}
\end{table}
\begin{table*}[]
\centering
\scalebox{0.80}{
\begin{tabular}{l|rrr|cccccccccc}
\toprule
 & & & & \multicolumn{2}{c}{LSP-Extended \cite{johnson+2011cvpr}} & \multicolumn{2}{c}{COCO \cite{lin+2014eccv}} & \multicolumn{2}{c}{PoseTrack \cite{andriluka+2018cvpr}} & \multicolumn{2}{c}{3DPW \cite{kolotouros+2019iccv}} & \multicolumn{2}{c}{Human3.6M \cite{ionescu+2014tpami}} \\
 & Param. & GFLOPs & FPS & P@0.05 & P@0.1 & P@0.05 & P@0.1 & P@0.05 & P@0.1 & M & PA-M & M & PA-M \\ \midrule
\rowcolor{gray!15}
HMR2.0 \cite{goel+2023iccv} & 670.5 & 125.6 & 62 & 53.3 & 82.4 & 86.1 & 96.2 & 89.7 & 97.7 & 81.3 & 54.3 & 49.7 & 32.1 \\
HMR2.0-L & 322.4 & 60.6 & 118 & 50.7 & 80.3 & 85.1 & 95.8 & 89.5 & 97.7 & 80.6 & 53.5 & 57.0 & 35.2 \\
HMR2.0-B & 92.8 & 17.3 & 366 & 42.8 & 74.0 & 82.1 & 94.6 & 85.9 & 96.2 & 80.0 & 52.7 & 61.3 & 38.5 \\
HMR2.0-S & 24.0 & 4.5 & 834 & 39.7 & 70.8 & 80.7 & 93.8 & 85.5 & 95.9 & 81.2 & 53.4 & 66.1 & 41.3 \\ \midrule
SwinHMR2.0-B-S4 & 95.5 & 18.0 & 265 & 45.4 & 76.5 & 82.4 & 94.7 & 86.2 & 96.3 & 82.2 & 54.6 & 51.5 & 34.3 \\
SwinHMR2.0-B-S3 & 66.1 & 16.2 & 289 & 43.9 & 75.4 & 82.6 & 94.6 & 86.7 & 96.2 & 81.1 & 54.0 & 52.1 & 34.8 \\
SwinHMR2.0-B-S2 & 9.3 & 3.1 & 881 & 24.8 & 53.5 & 67.0 & 87.2 & 73.3 & 90.6 & 94.2 & 59.1 & 69.1 & 48.3 \\
SwinHMR2.0-S-S4 & 52.3 & 10.2 & 377 & 43.6 & 75.7 & 82.9 & 94.6 & 86.9 & 96.4 & 81.3 & 54.4 & 52.1 & 34.8 \\
SwinHMR2.0-S-S3 & 35.8 & 9.1 & 408 & 41.0 & 73.7 & 82.0 & 94.4 & 86.3 & 96.0 & 80.9 & 53.3 & 52.6 & 35.2 \\
SwinHMR2.0-S-S2 & 3.8	& 1.7 & 1075 & 18.9 & 44.7 & 60.4 & 82.8 & 67.9 & 87.6 & 103.5 & 63.9 & 73.3 & 54.7 \\
SwinHMR2.0-T-S4 & 31.0 & 5.2 & 611 & 39.7 & 71.8 & 80.6 & 93.8 & 84.5 & 95.6 & 83.5 & 55.4 & 52.0 & 36.1 \\
SwinHMR2.0-T-S4 & 14.5 & 4.2 & 695 & 36.9 & 67.8 & 78.5 & 92.9 & 83.1 & 94.9 & 83.3 & 54.4 & 55.3 & 38.5 \\
SwinHMR2.0-T-S4 & 3.8 & 1.7 & 1076 & 20.2 & 46.3 & 61.1 & 83.5 & 68.6 & 87.9 & 100.3 & 62.0 & 73.9 & 54.1 \\ \midrule
GMFHMR2.0-B-S4 & 52.4 & 17.3 & 160 & 48.5 & 79.2 & 84.2 & 95.4 & 88.3 & 96.9 & 82.3 & 54.8 & 50.3 & 32.7 \\
GMFHMR2.0-B-S3 & 29.0 & 16.1 & 169 & 46.9 & 78.1 & 83.9 & 95.2 & 88.1 & 96.9 & 80.4 & 53.0 & 53.0 & 33.8 \\
GMFHMR2.0-B-S2 & 14.0 & 13.1 & 199 & 41.9 & 73.8 & 82.2 & 94.7 & 86.7 & 96.3 & 79.1 & 52.3 & 55.0 & 35.4 \\
GMFHMR2.0-S-S4 & 24.8 & 5.1 & 425 & 46.0 & 77.3 & 82.7 & 94.8 & 86.7 & 96.3 & 82.8 & 54.6 & 52.1 & 34.7 \\
GMFHMR2.0-S-S3 & 19.3 & 4.8 & 453 & 44.2 & 75.2 & 82.4 & 94.7 & 87.0 & 96.2 & 81.4 & 54.1 & 53.2 & 35.4 \\
GMFHMR2.0-S-S2 & 4.2 & 1.8 & 772 & 31.8 & 62.0 & 75.1 & 91.3 & 80.6 & 93.6 & 85.6 & 55.0 & 59.4 & 41.8 \\
GMFHMR2.0-T-S4 & 13.1 & 3.7 & 380 & 44.6 & 76.1 & 82.5 & 94.7 & 86.6 & 96.2 & 82.1 & 54.7 & 52.5 & 35.1 \\
GMFHMR2.0-T-S3 & 10.1 & 3.5 & 417 & 41.7 & 73.4 & 81.9 & 94.3 & 86.6 & 96.1 & 80.3 & 53.6 & 54.1 & 36.4 \\
GMFHMR2.0-T-S2 & 4.1 & 2.3 & 559 & 32.4 & 62.9 & 75.5 & 91.5 & 81.2 & 93.9 & 83.3 & 54.0 & 60.3 & 41.3 \\ \midrule
VMHMR2.0-B-S4 & 96.3 & 15.3 & 433 & 45.9 & 77.7 & 83.7 & 95.2 & 88.1 & 96.8 & 81.5 & 54.3 & 48.5 & 34.0 \\
VMHMR2.0-B-S3 & 62.1 & 13.6 & 487 & 44.6 & 76.1 & 82.9 & 95.0 & 87.0 & 96.5 & 80.7 & 53.6 & 51.6 & 34.0 \\
VMHMR2.0-B-S2 & 9.8 & 3.5 & 943 & 32.8 & 63.0 & 75.5 & 91.5 & 81.1 & 93.4 & 83.3 & 54.2 & 57.0 & 40.0 \\
VMHMR2.0-S-S4 & 52.9 & 8.6 & 540 & 46.1 & 77.1 & 83.4 & 94.9 & 87.8 & 96.6 & 81.4 & 54.3 & 53.0 & 34.8 \\
VMHMR2.0-S-S3 & 33.6 & 7.7 & 613 & 44.7 & 75.6 & 82.6 & 94.9 & 86.5 & 96.4 & 81.2 & 54.1 & 53.2 & 35.8 \\
VMHMR2.0-S-S2 & 4.1 & 2.0 & 1146 & 27.3 & 56.1 & 71.0 & 89.2 & 77.0 & 92.0 & 87.1 & 55.9 & 60.8 & 45.2 \\
VMHMR2.0-T-S4 & 33.0 & 4.9 & 751 & 43.5 & 74.6 & 82.5 & 94.5 & 86.6 & 96.3 & 82.0 & 54.1 & 51.8 & 34.8 \\
VMHMR2.0-T-S3 & 14.5	& 4.2 & 856 & 41.8 & 72.6 & 81.1 & 94.2 & 85.6 & 95.9 & 80.3 & 53.1 & 53.8 & 34.9 \\
VMHMR2.0-T-S2 & 3.8 & 1.6 & 1402 & 25.1 & 53.3 & 69.7 & 88.3 & 76.9 & 91.6 & 88.3 & 57.1 & 63.7 & 47.2 \\
\bottomrule
\end{tabular}
}
\caption{
2D and 3D pose estimation results of HMR models across five datasets.
In this table, ``P'' denotes PCK, ``M'' denotes MPJPE, and ``PA-M'' denotes PA-MPJPE.
Models shaded with gray cells are evaluated using publicly available pretrained weights, while the others are trained in our environment.
}
\label{tab:hmr:results:full}
\end{table*}

{
    \small
    \bibliographystyle{ieeenat_fullname}
    \bibliography{main}

@String(CVPR= {IEEE Conf. Comput. Vis. Pattern Recog.})

@String(ICCV= {Int. Conf. Comput. Vis.})

@String(ECCV= {Eur. Conf. Comput. Vis.})

@String(NIPS= {Adv. Neural Inform. Process. Syst.})

@String(TIP  = {IEEE Trans. Image Process.})

@String(TVCG  = {IEEE Trans. Vis. Comput. Graph.})

@String(ICME = {Int. Conf. Multimedia and Expo})

@String(ICLR = {Int. Conf. Learn. Represent.})

@String(IJCAI = {IJCAI})

@String(AAAI = {AAAI})

@String(CVPR  = {CVPR})

@String(ICCV  = {ICCV})

@String(ECCV  = {ECCV})

@String(NIPS  = {NeurIPS})

@String(TIP   = {IEEE TIP})

@String(TVCG  = {IEEE TVCG})

@String(ICME  =	{ICME})

@String(ICLR  = {ICLR})

@INPROCEEDINGS{johnson+2011cvpr,
  author={Johnson, Sam and Everingham, Mark},
  booktitle={CVPR}, 
  title="{Learning Effective Human Pose Estimation from Inaccurate Annotation}", 
  year={2011},
}

@INPROCEEDINGS{andriluka+2014cvpr,
  author={Andriluka, Mykhaylo and Pishchulin, Leonid and Gehler, Peter and Schiele, Bernt},
  booktitle={CVPR}, 
  title="{2D Human Pose Estimation: New Benchmark and State of the Art Analysis}", 
  year={2014},
}

@InProceedings{lin+2014eccv,
author={Lin, Tsung-Yi and Maire, Michael and Belongie, Serge and Hays, James and Perona, Pietro and Ramanan, Deva
and Doll{\'a}r, Piotr and Zitnick, C. Lawrence},
title="{Microsoft COCO: Common Objects in Context}",
booktitle={ECCV},
year={2014},
}

@ARTICLE{ionescu+2014tpami,
  author={Ionescu, Catalin and Papava, Dragos and Olaru, Vlad and Sminchisescu, Cristian},
  journal={TPAMI}, 
  title="{Human3.6M: Large Scale Datasets and Predictive Methods for 3D Human Sensing in Natural Environments}", 
  year={2014},
}

@article{loper+2015tog,
author = {Loper, Matthew and Mahmood, Naureen and Romero, Javier and Pons-Moll, Gerard and Black, Michael J.},
title = "{SMPL: A Skinned Multi-Person Linear Model}",
year = {2015},
journal = {ACM Trans. Graph.},
}

@INPROCEEDINGS{he+2016cvpr,
  author={He, Kaiming and Zhang, Xiangyu and Ren, Shaoqing and Sun, Jian},
  booktitle={CVPR}, 
  title="{Deep Residual Learning for Image Recognition}", 
  year={2016},
}

@inproceedings{bogo+eccv2016,
      title="{Keep it SMPL: Automatic Estimation of 3D Human Pose and Shape from a Single Image}", 
      author={Federica Bogo and Angjoo Kanazawa and Christoph Lassner and Peter Gehler and Javier Romero and Michael J. Black},
      year={2016},
booktitle={ECCV},
}

@INPROCEEDINGS{mehta+20173dv,
  author={Mehta, Dushyant and Rhodin, Helge and Casas, Dan and Fua, Pascal and Sotnychenko, Oleksandr and Xu, Weipeng and Theobalt, Christian},
  booktitle={3DV}, 
  title="{Monocular 3D Human Pose Estimation in the Wild Using Improved CNN Supervision}", 
  year={2017},
}

@inproceedings{vaswani+2017nips,
author = {Vaswani, Ashish and Shazeer, Noam and Parmar, Niki and Uszkoreit, Jakob and Jones, Llion and Gomez, Aidan N. and Kaiser, \L{}ukasz and Polosukhin, Illia},
title = "{Attention is All You Need}",
year = {2017},
booktitle = {NIPS},
}

@InProceedings{andriluka+2018cvpr,
author = {Andriluka, Mykhaylo and Iqbal, Umar and Insafutdinov, Eldar and Pishchulin, Leonid and Milan, Anton and Gall, Juergen and Schiele, Bernt},
title = "{PoseTrack: A Benchmark for Human Pose Estimation and Tracking}",
booktitle = {CVPR},
year = {2018}
}

@INPROCEEDINGS{gu+2018cvpr,
  author={Gu, Chunhui and Sun, Chen and Ross, David A. and Vondrick, Carl and Pantofaru, Caroline and Li, Yeqing and Vijayanarasimhan, Sudheendra and Toderici, George and Ricco, Susanna and Sukthankar, Rahul and Schmid, Cordelia and Malik, Jitendra},
  booktitle={CVPR}, 
  title="{AVA: A Video Dataset of Spatio-Temporally Localized Atomic Visual Actions}", 
  year={2018},
}

@inProceedings{kanazawa+2018cvpr,
  title="{End-to-end Recovery of Human Shape and Pose}",
  author = {Angjoo Kanazawa
  and Michael J. Black
  and David W. Jacobs
  and Jitendra Malik},
  booktitle={CVPR},
  year={2018}
}

@InProceedings{zanfir+2018cvpr,
author = {Zanfir, Andrei and Marinoiu, Elisabeta and Sminchisescu, Cristian},
title = "{Monocular 3D Pose and Shape Estimation of Multiple People in Natural Scenes - The Importance of Multiple Scene Constraints}",
booktitle = {CVPR},
year = {2018}
}

@inproceedings{zanfir+2018neurips,
    author = {Zanfir, Andrei and Marinoiu, Elisabeta and Zanfir, Mihai and Popa, Alin-Ionut and Sminchisescu, Cristian},
    title  = "{Deep Network for the Integrated 3D Sensing of Multiple People in Natural Images}",
    booktitle = {NeurIPS},
    year   = {2018}
}

@InProceedings{kanazawa+2019cvpr,
  title="{Learning 3D Human Dynamics from Video}",
  author = {Angjoo Kanazawa and Jason Y. Zhang and Panna Felsen and Jitendra Malik},
  booktitle={CVPR},
  year={2019}
}

@inproceedings{sun+2019cvpr,
  title="{Deep High-Resolution Representation Learning for Human Pose Estimation}",
  author={Sun, Ke and Xiao, Bin and Liu, Dong and Wang, Jingdong},
  booktitle={CVPR},
  year={2019}
}

@inproceedings{kolotouros+2019iccv,
      title="{Learning to Reconstruct 3D Human Pose and Shape via Model-fitting in the Loop}", 
      author={Nikos Kolotouros and Georgios Pavlakos and Michael J. Black and Kostas Daniilidis},
      year={2019},
booktitle={ICCV},
}

@inproceedings{pavlakos+2019cvpr,
      title="{Expressive Body Capture: 3D Hands, Face, and Body from a Single Image}", 
      author={Georgios Pavlakos and Vasileios Choutas and Nima Ghorbani and Timo Bolkart and Ahmed A. A. Osman and Dimitrios Tzionas and Michael J. Black},
      year={2019},
booktitle={CVPR},
}

@INPROCEEDINGS{wu+2019icme,
  author={Wu, Jiahong and Zheng, He and Zhao, Bo and Li, Yixin and Yan, Baoming and Liang, Rui and Wang, Wenjia and Zhou, Shipei and Lin, Guosen and Fu, Yanwei and Wang, Yizhou and Wang, Yonggang},
  booktitle={ICME}, 
  title="{Large-Scale Datasets for Going Deeper in Image Understanding}", 
  year={2019},
}

@InProceedings{tan+2019icml,
  title = 	 "{EfficientNet: Rethinking Model Scaling for Convolutional Neural Networks}",
  author =       {Tan, Mingxing and Le, Quoc},
  booktitle = 	 {ICML},
  year = 	 {2019},
}

@inproceedings{kocabas+2020cvpr,
      title="{VIBE: Video Inference for Human Body Pose and Shape Estimation}", 
      author={Muhammed Kocabas and Nikos Athanasiou and Michael J. Black},
      year={2020},
booktitle={CVPR},
}

@inproceedings{choi+2021cvpr,
      title="{Beyond Static Features for Temporally Consistent 3D Human Pose and Shape from a Video}", 
      author={Hongsuk Choi and Gyeongsik Moon and Ju Yong Chang and Kyoung Mu Lee},
      year={2021},
booktitle={CVPR},
}

@inproceedings{lin+2021cvpr,
author = {Lin, Kevin and Wang, Lijuan and Liu, Zicheng},
title = "{End-to-End Human Pose and Mesh Reconstruction with Transformers}",
booktitle = {CVPR},
year = {2021},
}

@InProceedings{xu+2021cvpr,
    author    = {Xu, Lumin and Guan, Yingda and Jin, Sheng and Liu, Wentao and Qian, Chen and Luo, Ping and Ouyang, Wanli and Wang, Xiaogang},
    title     = "{ViPNAS: Efficient Video Pose Estimation via Neural Architecture Search}",
    booktitle = {CVPR},
    year      = {2021},
}

@INPROCEEDINGS{yu+2021cvpr,
  author={Yu, Changqian and Xiao, Bin and Gao, Changxin and Yuan, Lu and Zhang, Lei and Sang, Nong and Wang, Jingdong},
  booktitle={CVPR}, 
  title="{Lite-HRNet: A Lightweight High-Resolution Network}", 
  year={2021},
}

@InProceedings{fan+2021iccv,
    author    = {Fan, Haoqi and Xiong, Bo and Mangalam, Karttikeya and Li, Yanghao and Yan, Zhicheng and Malik, Jitendra and Feichtenhofer, Christoph},
    title     = "{Multiscale Vision Transformers}",
    booktitle = {ICCV},
    year      = {2021},
}

@inproceedings{kocabas+2021iccv,
  title = "{PARE: Part Attention Regressor for {3D} Human Body Estimation}",
  author = {Kocabas, Muhammed and Huang, Chun-Hao P. and Hilliges, Otmar and Black, Michael J.},
  booktitle = {ICCV},
  year = {2021},
}

@Inproceedings{kolotouros+2021iccv,
  Title          = "{Probabilistic Modeling for Human Mesh Recovery}",
  Author         = {Kolotouros, Nikos and Pavlakos, Georgios and Jayaraman, Dinesh and Daniilidis, Kostas},
  Booktitle      = {ICCV},
  Year           = {2021}
}

@inproceedings{lin+2021iccv,
author = {Lin, Kevin and Wang, Lijuan and Liu, Zicheng},
title = "{Mesh Graphormer}",
booktitle = {ICCV},
year = {2021},
}

@InProceedings{liu+2021iccv,
    author    = {Liu, Ze and Lin, Yutong and Cao, Yue and Hu, Han and Wei, Yixuan and Zhang, Zheng and Lin, Stephen and Guo, Baining},
    title     = "{Swin Transformer: Hierarchical Vision Transformer using Shifted Windows}",
    booktitle = {ICCV},
    year      = {2021},
}

@InProceedings{sun+2021iccv,
    author = {Sun, Yu and Bao, Qian and Liu, Wu and Fu, Yili and Michael J., Black and Mei, Tao},
    title = "{Monocular, One-stage, Regression of Multiple 3D People}",
    booktitle = {ICCV},
    year = {2021}
}

@InProceedings{wang+2021iccv,
    author    = {Wang, Wenhai and Xie, Enze and Li, Xiang and Fan, Deng-Ping and Song, Kaitao and Liang, Ding and Lu, Tong and Luo, Ping and Shao, Ling},
    title     = "{Pyramid Vision Transformer: A Versatile Backbone for Dense Prediction Without Convolutions}",
    booktitle = {ICCV},
    year      = {2021},
}

@inproceedings{zhang+2021iccv,
  title="{PyMAF: 3D Human Pose and Shape Regression with Pyramidal Mesh Alignment Feedback Loop}",
  author={Zhang, Hongwen and Tian, Yating and Zhou, Xinchi and Ouyang, Wanli and Liu, Yebin and Wang, Limin and Sun, Zhenan},
  booktitle={ICCV},
  year={2021}
}

@article{dosovitskiy+2020iclr,
  title="{An Image is Worth 16x16 Words: Transformers for Image Recognition at Scale}",
  author={Dosovitskiy, Alexey and Beyer, Lucas and Kolesnikov, Alexander and Weissenborn, Dirk and Zhai, Xiaohua and Unterthiner, Thomas and  Dehghani, Mostafa and Minderer, Matthias and Heigold, Georg and Gelly, Sylvain and Uszkoreit, Jakob and Houlsby, Neil},
  journal={ICLR},
  year={2021}
}

@INPROCEEDINGS{luo+2021icme,
  author={Luo, Zhengxion and Wang, Zhicheng and Cai, Yuanhao and Wang, Guanan and Wang, Liang and Huang, Yan and Zhou, ErJin and Tan, Tieniu and Sun, Jian},
  booktitle={ICME}, 
  title="{Efficient Human Pose Estimation by Learning Deeply Aggregated Representations}", 
  year={2021},
}

@InProceedings{radford+2021icml,
  title = 	 "{Learning Transferable Visual Models from Natural Language Supervision}",
  author =       {Radford, Alec and Kim, Jong Wook and Hallacy, Chris and Ramesh, Aditya and Goh, Gabriel and Agarwal, Sandhini and Sastry, Girish and Askell, Amanda and Mishkin, Pamela and Clark, Jack and Krueger, Gretchen and Sutskever, Ilya},
  booktitle = 	 {ICML},
  year = 	 {2021},
}

@InProceedings{touvron+2021icml,
  title = 	 "{Training Data-efficient Image Transformers \& Distillation through Attention}",
  author =       {Touvron, Hugo and Cord, Matthieu and Douze, Matthijs and Massa, Francisco and Sablayrolles, Alexandre and J\'egou, Herv\'e},
  booktitle = 	 {ICML},
  year = 	 {2021},
}

@inproceedings{chu+2021neurips,
      title="{Twins: Revisiting the Design of Spatial Attention in Vision Transformers}", 
      author={Xiangxiang Chu and Zhi Tian and Yuqing Wang and Bo Zhang and Haibing Ren and Xiaolin Wei and Huaxia Xia and Chunhua Shen},
      year={2021},
booktitle={NeurIPS},
}

@inproceedings{xie+2021neurips,
      title="{SegFormer: Simple and Efficient Design for Semantic Segmentation with Transformers}", 
      author={Enze Xie and Wenhai Wang and Zhiding Yu and Anima Anandkumar and Jose M. Alvarez and Ping Luo},
      year={2021},
booktitle={NeurIPS},
}

@ARTICLE{wang+2021tpami,
  author={Wang, Jingdong and Sun, Ke and Cheng, Tianheng and Jiang, Borui and Deng, Chaorui and Zhao, Yang and Liu, Dong and Mu, Yadong and Tan, Mingkui and Wang, Xinggang and Liu, Wenyu and Xiao, Bin},
  journal={TPAMI}, 
  title="{Deep High-Resolution Representation Learning for Visual Recognition}", 
  year={2021},
}

@INPROCEEDINGS{li+2021wacv,
  author={Li, Zhongguo and Oskarsson, Magnus and Heyden, Anders},
  booktitle={WACV}, 
  title="{3D Human Pose and Shape Estimation Through Collaborative Learning and Multi-view Model-fitting}", 
  year={2021},
}

@article{wang+2022cvmj,
  title="{PVT v2: Improved Baselines with Pyramid Vision Transformer}",
  author={Wang, Wenhai and Xie, Enze and Li, Xiang and Fan, Deng-Ping and Song, Kaitao and Liang, Ding and Lu, Tong and Luo, Ping and Shao, Ling},
  journal={CVMJ},
  year={2022},
}

@InProceedings{choi+2022cvpr,  
author = {Choi, Hongsuk and Moon, Gyeongsik and Park, JoonKyu and Lee, Kyoung Mu},  
title = "{Learning to Estimate Robust 3D Human Mesh from In-the-Wild Crowded Scenes}",  
booktitle = {CVPR},
year = {2022},
}

@InProceedings{li+2022cvpr,
    author    = {Li, Yanghao and Wu, Chao-Yuan and Fan, Haoqi and Mangalam, Karttikeya and Xiong, Bo and Malik, Jitendra and Feichtenhofer, Christoph},
    title     = "{MViTv2: Improved Multiscale Vision Transformers for Classification and Detection}",
    booktitle = {CVPR},
    year      = {2022},
}

@inproceedings{liu+2022cvpr,
  title="{Swin Transformer V2: Scaling Up Capacity and Resolution}", 
  author={Ze Liu and Han Hu and Yutong Lin and Zhuliang Yao and Zhenda Xie and Yixuan Wei and Jia Ning and Yue Cao and Zheng Zhang and Li Dong and Furu Wei and Baining Guo},
  booktitle={CVPR},
  year={2022}
}

@inproceedings{pavlakos+2022cvpr,
    title = "{Human Mesh Recovery from Multiple Shots}",
    author = {Pavlakos, Georgios and Malik, Jitendra and Kanazawa, Angjoo},
    booktitle={CVPR},
    year = {2022}
}

@InProceedings{sun+2022cvpr,
    author    = {Sun, Yu and Liu, Wu and Bao, Qian and Fu, Yili and Mei, Tao and Black, Michael J.},
    title     = "{Putting People in Their Place: Monocular Regression of 3D People in Depth}",
    booktitle = {CVPR},
    year      = {2022},
}

@InProceedings{wang+2022cvpr,
    author    = {Wang, Yihan and Li, Muyang and Cai, Han and Chen, Wei-Ming and Han, Song},
    title     = "{Lite Pose: Efficient Architecture Design for 2D Human Pose Estimation}",
    booktitle = {CVPR},
    year      = {2022},
}

@inproceedings{wei+2022cvpr,
      title="{Capturing Humans in Motion: Temporal-Attentive 3D Human Pose and Shape Estimation from Monocular Video}", 
      author={Wen-Li Wei and Jen-Chun Lin and Tyng-Luh Liu and Hong-Yuan Mark Liao},
      year={2022},
booktitle={CVPR},
}

@InProceedings{cho+2022eccv,
    title="{Cross-Attention of Disentangled Modalities for 3D Human Mesh Recovery with Transformers}",
    author={Junhyeong Cho and Kim Youwang and Tae-Hyun Oh},
    booktitle={ECCV},
    year={2022}
}

@inproceedings{li+2022eccv,
      title="{SimCC: a Simple Coordinate Classification Perspective for Human Pose Estimation}", 
      author={Yanjie Li and Sen Yang and Peidong Liu and Shoukui Zhang and Yunxiao Wang and Zhicheng Wang and Wankou Yang and Shu-Tao Xia},
      year={2022},
booktitle={ECCV},
}

@Inproceedings{li+2022eccv2,
  Title     = "{CLIFF: Carrying Location Information in Full Frames into Human Pose and Shape Estimation}",
  Author    = {Li, Zhihao and Liu, Jianzhuang and Zhang, Zhensong and Xu, Songcen and Yan, Youliang},
  Booktitle = {ECCV},
  Year      = {2022}
}

@inproceedings{cao+2022ijcai,
   title="{AggPose: Deep Aggregation Vision Transformer for Infant Pose Estimation}",
   booktitle={IJCAI},
   author={Cao, Xu and Li, Xiaoye and Ma, Liya and Huang, Yi and Feng, Xuan and Chen, Zening and Zeng, Hongwu and Cao, Jianguo},
   year={2022},
}

@inproceedings{pang+2022neurips,
      title="{Benchmarking and Analyzing 3D Human Pose and Shape Estimation beyond Algorithms}", 
      author={Hui En Pang and Zhongang Cai and Lei Yang and Tianwei Zhang and Ziwei Liu},
      year={2022},
booktitle={NeurIPS},
}

@inproceedings{yu+2022neurips,
 author = {Yu, Zhixuan and Zhang, Linguang and Xu, Yuanlu and Tang, Chengcheng and TRAN, LUAN and Keskin, Cem and Park, Hyun Soo},
 booktitle = {NeurIPS},
 title = "{Multiview Human Body Reconstruction from Uncalibrated Cameras}",
 year = {2022},
}

@inproceedings{xu+2022neurips,
 author = {Xu, Yufei and Zhang, Jing and ZHANG, Qiming and Tao, Dacheng},
 booktitle = {NeurIPS},
 title = "{ViTPose: Simple Vision Transformer Baselines for Human Pose Estimation}",
 year = {2022},
}

@article{yang+2022pr,
title = "{Searching Part-specific Neural Fabrics for Human Pose Estimation}",
journal = {Pattern Recognition},
year = {2022},
author = {Sen Yang and Wankou Yang and Zhen Cui},
}

@article{li+2022sensors,
  author    = {Jie Li and
               Zhixing Wang and
               Bo Qi and
               Jianlin Zhang and
               Hu Yang},
  title     = "{MEMe: {A} Mutually Enhanced Modeling Method for Efficient and Effective Human Pose Estimation}",
  journal   = {Sensors},
  year      = {2022},
}

@article{steiner+2022tmlr,
title="{How to train your ViT? Data, Augmentation, and Regularization in Vision Transformers}",
author={Andreas Peter Steiner and Alexander Kolesnikov and Xiaohua Zhai and Ross Wightman and Jakob Uszkoreit and Lucas Beyer},
journal={TMLR},
year={2022},
}

@article{ge+2023arxiv,
      title="{Advancing Vision Transformers with Group-Mix Attention}", 
      author={Chongjian Ge and Xiaohan Ding and Zhan Tong and Li Yuan and Jiangliu Wang and Yibing Song and Ping Luo},
      year={2023},
      journal={arXiv preprint arxiv:2311.15157},
}

@InProceedings{jia+2023aaai,
title="{Delving Deep into Pixel Alignment Feature for Accurate Multi-view Human Mesh Recovery}",
author={Jia, Kai and Zhang, Hongwen and An, Liang and Liu, Yebin},
booktitle={AAAI},
year={2023},
}

@InProceedings{beyer+2023cvpr,
    author    = {Beyer, Lucas and Izmailov, Pavel and Kolesnikov, Alexander and Caron, Mathilde and Kornblith, Simon and Zhai, Xiaohua and Minderer, Matthias and Tschannen, Michael and Alabdulmohsin, Ibrahim and Pavetic, Filip},
    title     = "{FlexiViT: One Model for All Patch Sizes}",
    booktitle = {CVPR},
    year      = {2023},
}

@InProceedings{kim+2023cvpr,
author = {Kim, Jeonghwan and Gwon, Mi-Gyeong and Park, Hyunwoo and Kwon, Hyukmin and Um, Gi-Mun and Kim, Wonjun},
title = "{Sampling is Matter: Point-guided 3D Human Mesh Reconstruction}",
booktitle = {CVPR},
year = {2023}
}

@INPROCEEDINGS{lin+2023cvpr,
  author={Lin, Jing and Zeng, Ailing and Wang, Haoqian and Zhang, Lei and Li, Yu},
  booktitle={CVPR}, 
  title="{One-Stage 3D Whole-Body Mesh Recovery with Component Aware Transformer}", 
  year={2023},
}

@InProceedings{ma+2023cvpr,
    author    = {Ma, Xiaoxuan and Su, Jiajun and Wang, Chunyu and Zhu, Wentao and Wang, Yizhou},
    title     = "{3D Human Mesh Estimation From Virtual Markers}",
    booktitle = {CVPR},
    year      = {2023},
}

@inproceedings{qiu+2023cvpr,
      title="{PSVT: End-to-End Multi-person 3D Pose and Shape Estimation with Progressive Video Transformers}", 
      author={Zhongwei Qiu and Yang Qiansheng and Jian Wang and Haocheng Feng and Junyu Han and Errui Ding and Chang Xu and Dongmei Fu and Jingdong Wang},
      year={2023},
booktitle={CVPR},
}

@InProceedings{tian+2023cvpr,
    author    = {Tian, Rui and Wu, Zuxuan and Dai, Qi and Hu, Han and Qiao, Yu and Jiang, Yu-Gang},
    title     = "{ResFormer: Scaling ViTs With Multi-Resolution Training}",
    booktitle = {CVPR},
    year      = {2023},
}

@INPROCEEDINGS{yoshiyasu+2023cvpr,
  author={Yoshiyasu, Yusuke},
  booktitle={CVPR}, 
  title="{Deformable Mesh Transformer for 3D Human Mesh Recovery}", 
  year={2023},
}

@inproceedings{zheng+2023cvpr,
    title="{POTTER: Pooling Attention Transformer for Efficient Human Mesh Recovery}",
    author={Zheng, Ce and Liu, Xianpeng and Qi, Guo-Jun and Chen, Chen},
    booktitle={CVPR},
    year={2023}
}

@InProceedings{dou+2023iccv,
    author    = {Dou, Zhiyang and Wu, Qingxuan and Lin, Cheng and Cao, Zeyu and Wu, Qiangqiang and Wan, Weilin and Komura, Taku and Wang, Wenping},
    title     = "{TORE: Token Reduction for Efficient Human Mesh Recovery with Transformer}",
    booktitle = {ICCV},
    year      = {2023},
}

@InProceedings{goel+2023iccv,
    author    = {Goel, Shubham and Pavlakos, Georgios and Rajasegaran, Jathushan and Kanazawa, Angjoo and Malik, Jitendra},
    title     = "{Humans in 4D: Reconstructing and Tracking Humans with Transformers}",
    booktitle = {ICCV},
    year      = {2023},
}

@InProceedings{wei+2023iccv,
    author    = {Wei, Yixuan and Hu, Han and Xie, Zhenda and Liu, Ze and Zhang, Zheng and Cao, Yue and Bao, Jianmin and Chen, Dong and Guo, Baining},
    title     = "{Improving CLIP Fine-tuning Performance}",
    booktitle = {ICCV},
    year      = {2023},
}

@inproceedings{zhang+2023iccv,
  title = "{Probabilistic Human Mesh Recovery in 3D Scenes from Egocentric Views}",
  author = {Siwei Zhang and Qianli Ma and Yan Zhang and Sadegh Aliakbarian and Darren Cosker and Siyu Tang},
  booktitle = {ICCV},
  year = {2023}
}

@inproceedings{yang+2023iclr,
title="{Capturing the Motion of Every Joint: 3D Human Pose and Shape Estimation with Independent Tokens}",
author={Sen Yang and Wen Heng and Gang Liu and GUOZHONG LUO and Wankou Yang and Gang YU},
booktitle={ICLR},
year={2023},
}

@InProceedings{yang+2023iclr2,
      title="{GPViT: A High Resolution Non-Hierarchical Vision Transformer with Group Propagation}", 
      author={Chenhongyi Yang and Jiarui Xu and Shalini De Mello and Elliot J. Crowley and Xiaolong Wang},
      booktitle={ICLR},
      year={2023},
}

@inproceedings{zhang+2023iclr,
  title="{HiViT: A Simpler and More Efficient Design of Hierarchical Vision Transformer}",
  author={Zhang, Xiaosong and Tian, Yunjie and Xie, Lingxi and Huang, Wei and Dai, Qi and Ye, Qixiang and Tian, Qi},
  booktitle={ICLR},
  year={2023},
}

@inproceedings{cai+2023neurips,
    title="{SMPLer-X: Scaling Up Expressive Human Pose and Shape Estimation}",
    author={Cai, Zhongang and Yin, Wanqi and Zeng, Ailing and Wei, Chen and Sun, Qingping and Yanjun, Wang and Pang, Hui En and Mei, Haiyi and Zhang, Mingyuan and Zhang, Lei and Loy, Chen Change and Yang, Lei and Liu, Ziwei},
    booktitle={NeurIPS},
    year={2023}
}

@article{zheng+2023acm,
author = {Zheng, Ce and Wu, Wenhan and Chen, Chen and Yang, Taojiannan and Zhu, Sijie and Shen, Ju and Kehtarnavaz, Nasser and Shah, Mubarak},
title = "{Deep Learning-based Human Pose Estimation: A Survey}",
year = {2023},
journal = {ACM Comput. Surv.},
}

@ARTICLE{sun+2023tpami,
  author={Sun, Weixuan and Qin, Zhen and Deng, Hui and Wang, Jianyuan and Zhang, Yi and Zhang, Kaihao and Barnes, Nick and Birchfield, Stan and Kong, Lingpeng and Zhong, Yiran},
  journal={TPAMI}, 
  title="{Vicinity Vision Transformer}", 
  year={2023},
}

@ARTICLE{tian+2023tpami,
  author={Tian, Yating and Zhang, Hongwen and Liu, Yebin and Wang, Limin},
  journal={TPAMI}, 
  title="{Recovering 3D Human Mesh From Monocular Images: A Survey}", 
  year={2023},
}

@article{zhang+2023tpami,
  title="{PyMAF-X: Towards Well-aligned Full-body Model Regression from Monocular Images}",
  author={Zhang, Hongwen and Tian, Yating and Zhang, Yuxiang and Li, Mengcheng and An, Liang and Sun, Zhenan and Liu, Yebin},
  journal={TPAMI},
  year={2023}
}

@inproceedings{shen+2024acmmm,
author = {Shen, Wenhao and Yin, Wanqi and Wang, Hao and Wei, Chen and Cai, Zhongang and Yang, Lei and Lin, Guosheng},
title = "{HMR-Adapter: A Lightweight Adapter with Dual-Path Cross Augmentation for Expressive Human Mesh Recovery}",
year = {2024},
booktitle = {ACM MM}
}

@InProceedings{dwivedi+2024cvpr,
    author    = {Dwivedi, Sai Kumar and Sun, Yu and Patel, Priyanka and Feng, Yao and Black, Michael J.},
    title     = "{TokenHMR: Advancing Human Mesh Recovery with a Tokenized Pose Representation}",
    booktitle = {CVPR},
    year      = {2024},
}

@InProceedings{feng+2024cvpr,
    author    = {Feng, Yao and Lin, Jing and Dwivedi, Sai Kumar and Sun, Yu and Patel, Priyanka and Black, Michael J.},
    title     = "{ChatPose: Chatting about 3D Human Pose}",
    booktitle = {CVPR},
    year      = {2024},
}

@InProceedings{ge+2024cvpr,
    author    = {Ge, Haoyang and Feng, Qiao and Jia, Hailong and Li, Xiongzheng and Yin, Xiangjun and Zhou, You and Yang, Jingyu and Li, Kun},
    title     = "{LPSNet: End-to-End Human Pose and Shape Estimation with Lensless Imaging}",
    booktitle = {CVPR},
    year      = {2024},
}

@InProceedings{le+2024cvpr,
    author    = {Le, Eric-Tuan and Kakolyris, Antonis and Koutras, Petros and Tam, Himmy and Skordos, Efstratios and Papandreou, George and G\"uler, Riza Alp and Kokkinos, Iasonas},
    title     = "{MeshPose: Unifying DensePose and 3D Body Mesh Reconstruction}",
    booktitle = {CVPR},
    year      = {2024},
}

@InProceedings{stathopoulos+2024cvpr,
    author    = {Stathopoulos, Anastasis and Han, Ligong and Metaxas, Dimitris},
    title     = "{Score-Guided Diffusion for 3D Human Recovery}",
    booktitle = {CVPR},
    year      = {2024},
}

@InProceedings{sun+2024cvpr,
    author    = {Sun, Qingping and Wang, Yanjun and Zeng, Ailing and Yin, Wanqi and Wei, Chen and Wang, Wenjia and Mei, Haiyi and Leung, Chi-Sing and Liu, Ziwei and Yang, Lei and Cai, Zhongang},
    title     = "{AiOS: All-in-One-Stage Expressive Human Pose and Shape Estimation}",
    booktitle = {CVPR},
    year      = {2024},
}

@InProceedings{xu+2024cvpr,
    author    = {Xu, Yuan and Ma, Xiaoxuan and Su, Jiajun and Zhu, Wentao and Yu, Qiao and Wang, Yizhou},
    title     = "{ScoreHypo: Probabilistic Human Mesh Estimation with Hypothesis Scoring}",
    booktitle = {CVPR},
    year      = {2024}
}

@InProceedings{agarwal+2024eccvw,
author={Agarwal, Vatsal
and Levy, Mara
and Ehrlich, Max
and Tang, Youbao
and Zhang, Ning
and Shrivastava, Abhinav},
title="{Coarse-to-Fine Human Mesh Recovery with Transformers}",
booktitle={ECCV Workshops},
year={2025},
}

@inproceedings{fabien+2024eccv,
    title="{Multi-HMR: Multi-Person Whole-Body Human Mesh Recovery in a Single Shot}",
    author={Baradel, Fabien and 
            Armando, Matthieu and 
            Galaaoui, Salma and 
            Br{\'e}gier, Romain and 
            Weinzaepfel, Philippe and 
            Rogez, Gr{\'e}gory and
            Lucas, Thomas
            },
    booktitle={ECCV},
    year={2024}
}

@inproceedings{fiche+2024eccv,
      title="{VQ-HPS: Human Pose and Shape Estimation in a Vector-Quantized Latent Space}", 
      author={Guénolé Fiche and Simon Leglaive and Xavier Alameda-Pineda and Antonio Agudo and Francesc Moreno-Noguer},
      year={2024},
booktitle={ECCV},
}

@InProceedings{xiao+2024eccv,
author={Xiao, Yabo and He, Mingshu and Yu, Dongdong},
title="{Global-to-Pixel Regression for Human Mesh Recovery}",
booktitle={ECCV},
year={2024},
}

@inproceedings{xie+2024icmr,
author = {Xie, Zhenyu and He, Huanyu and Zou, Gui and Wu, Jie and Liu, Guoliang and Zhao, Jun and Wang, Yingxue and Lin, Hui and Lin, Weiyao},
title = "{Visibility-guided Human Body Reconstruction from Uncalibrated Multi-view Cameras}",
year = {2024},
booktitle = {ICMR},
}

@article{fang+2024ivc,
   title="{EVA-02: A Visual Representation for Neon Genesis}",
   journal={Image and Vision Computing},
   author={Fang, Yuxin and Sun, Quan and Wang, Xinggang and Huang, Tiejun and Wang, Xinlong and Cao, Yue},
   year={2024},
 }

@inproceedings{liu+2024neurips,
      title="{VMamba: Visual State Space Model}", 
      author={Yue Liu and Yunjie Tian and Yuzhong Zhao and Hongtian Yu and Lingxi Xie and Yaowei Wang and Qixiang Ye and Yunfan Liu},
      year={2024},
      booktitle={NeurIPS},
}

@inproceedings{sarandi+2024neurips,
 author = {S\'{a}r\'{a}ndi, Istv\'{a}n and Pons-Moll, Gerard},
 booktitle = {NeurIPS},
 title = "{Neural Localizer Fields for Continuous 3D Human Pose and Shape Estimation}",
 year = {2024}
}

@article{liu+2024neuro,
title = "{Deep learning for 3D Human Pose Estimation and Mesh Recovery: A Survey}",
journal = {Neurocomputing},
year = {2024},
author = {Yang Liu and Changzhen Qiu and Zhiyong Zhang},
}

@inproceedings{shen+2024sa,
  title="{World-Grounded Human Motion Recovery via Gravity-View Coordinates}",
  author={Shen, Zehong and Pi, Huaijin and Xia, Yan and Cen, Zhi and Peng, Sida and Hu, Zechen and Bao, Hujun and Hu, Ruizhen and Zhou, Xiaowei},
  booktitle={SIGGRAPH Asia},
  year={2024}
}

@ARTICLE{li+2024tip,
  author={Li, Wenhao and Liu, Mengyuan and Liu, Hong and Ren, Bin and Li, Xia and You, Yingxuan and Sebe, Nicu},
  journal={TIP}, 
  title="{HYRE: Hybrid Regressor for 3D Human Pose and Shape Estimation}", 
  year={2025},
}

@ARTICLE{cai+2024tpami,
            author={Cai, Zhongang and Zhang, Mingyuan and Ren, Jiawei and Wei, Chen and Ren, Daxuan and 
                    Lin, Zhengyu and Zhao, Haiyu and Yang, Lei and Loy, Chen Change and Liu, Ziwei},
            journal={TPAMI}, 
            title="{Playing for 3D Human Recovery}", 
            year={2024},
        }

@Article{gao+2025air,
author={Gao, Zheyan and Chen, Jinyan and Liu, Yuxin and Jin, Yucheng and Tian, Dingxiaofei},
title="{A Systematic Survey on Human Pose Estimation: Upstream and Downstream Tasks, Approaches, Lightweight Models, and Prospects}",
journal={Artificial Intelligence Review},
year={2025},
}

@article{saleem+2025aaai, 
title="{GenHMR: Generative Human Mesh Recovery}", 
journal={AAAI}, 
author={Saleem, Muhammad Usama and Pinyoanuntapong, Ekkasit and Wang, Pu and Xue, Hongfei and Das, Srijan and Chen, Chen}, 
year={2025},
}

@article{zhu+2025aaai,
  title="{MUC: Mixture of Uncalibrated Cameras for Robust 3D Human Body Reconstruction}",
  author={Zhu, Yitao and Wang, Sheng and Xu, Mengjie and Zhuang, Zixu and Wang, Zhixin and Wang, Kaidong and Zhang, Han and Wang, Qian},
journal={AAAI}, 
  year={2025},
}

@InProceedings{fiche+2025cvpr,
    author    = {Fiche, Gu\'enol\'e and Leglaive, Simon and Alameda-Pineda, Xavier and Moreno-Noguer, Francesc},
    title     = "{MEGA: Masked Generative Autoencoder for Human Mesh Recovery}",
    booktitle = {CVPR},
    year      = {2025},
}

@InProceedings{lin+2025cvpr,
    author    = {Lin, Jing and Feng, Yao and Liu, Weiyang and Black, Michael J.},
    title     = "{ChatHuman: Chatting about 3D Humans with Tools}",
    booktitle = {CVPR},
    year      = {2025},
}

@InProceedings{wang+2025cvpr,
    author    = {Wang, Yufu and Sun, Yu and Patel, Priyanka and Daniilidis, Kostas and Black, Michael J. and Kocabas, Muhammed},
    title     = "{PromptHMR: Promptable Human Mesh Recovery}",
    booktitle = {CVPR},
    year      = {2025},
}

@InProceedings{wu+2025cvpr,
    author    = {Wu, Ziyu and Xiong, Yufan and Niu, Mengting and Xie, Fangting and Wan, Quan and Ying, Qijun and Liu, Boyan and Cai, Xiaohui},
    title     = "{PI-HMR: Towards Robust In-bed Temporal Human Shape Reconstruction with Contact Pressure Sensing}",
    booktitle = {CVPR},
    year      = {2025},
}

@InProceedings{su+2025cvpr,
    author    = {Su, Chi and Ma, Xiaoxuan and Su, Jiajun and Wang, Yizhou},
    title     = "{SAT-HMR: Real-Time Multi-Person 3D Mesh Estimation via Scale-Adaptive Tokens}",
    booktitle = {CVPR},
    year      = {2025},
}

@InProceedings{zhang+2025cvpr,
    author    = {Zhang, Yuhong and Wu, Guanlin and Chen, Ling-Hao and Zhao, Zhuokai and Lin, Jing and Jiang, Xiaoke and Wu, Jiamin and Li, Zhuoheng and Yang, Hao Frank and Wang, Haoqian and Zhang, Lei},
    title     = "{HumanMM: Global Human Motion Recovery from Multi-shot Videos}",
    booktitle = {CVPR},
    year      = {2025},
}

@InProceedings{kang+2025cvprw,
    author    = {Kang, Xiyuan and Yuan, Yi and Dong, Xu and Awais, Muhammad and Tang, Lilian and Kittler, Josef and Feng, Zhenhua},
    title     = "{Short-term 3D Human Mesh Recovery with Virtual Markers Disentanglement}",
    booktitle = {CVPR Workshops},
    year      = {2025},
}

@InProceedings{prospero+2025cvprw,
    author    = {Prospero, Lorenza and Hamdi, Abdullah and Henriques, Joao F. and Rupprecht, Christian},
    title     = "{GST: Precise 3D Human Body from a Single Image with Gaussian Splatting Transformers}",
    booktitle = {CVPR Workshops},
    year      = {2025},
}

@inproceedings{romain+2025cvprw,
      title="{CondiMen: Conditional Multi-Person Mesh Recovery}", 
      author={Brégier Romain and Baradel Fabien and Lucas Thomas and Galaaoui Salma and Armando Matthieu and Weinzaepfel Philippe and Rogez Grégory},
      year={2025},
booktitle={CVPR Workshops}
}

@misc{alkin+2025iclr,
      title="{Vision-LSTM: xLSTM as Generic Vision Backbone}", 
      author={Benedikt Alkin and Maximilian Beck and Korbinian Pöppel and Sepp Hochreiter and Johannes Brandstetter},
      year={2025},
booktitle={ICLR},
}

@inproceedings{shen+2025icml,
  title="{ADHMR: Aligning Diffusion-based Human Mesh Recovery via Direct Preference Optimization}",
  author={Shen, Wenhao and Yin, Wanqi and Yang, Xiaofeng and Chen, Cheng and Song, Chaoyue and Cai, Zhongang and Yang, Lei and Wang, Hao and Lin, Guosheng},
  booktitle={ICML},
  year={2025},
}

@ARTICLE{li+2025tpami,
  author={Li, Jiefeng and Bian, Siyuan and Xu, Chao and Chen, Zhicun and Yang, Lixin and Lu, Cewu},
  journal={TPAMI}, 
  title="{HybrIK-X: Hybrid Analytical-Neural Inverse Kinematics for Whole-Body Mesh Recovery}", 
  year={2025},
}

@ARTICLE{liao+2024tvcg,
  author={Liao, Xinyao and Zhang, Chen and Xu, Jianyao and Su, Wanjuan and Chen, Zhi and Tao, Wenbing},
  journal={TVCG}, 
  title="{InstaHMR: Instance-Aware One-Stage Multi-Person Human Mesh Recovery}", 
  year={2025},
}

@inproceedings{patel+20243dv,
      title="{CameraHMR: Aligning People with Perspective}", 
      author={Priyanka Patel and Michael J. Black},
      year={2024},
booktitle={3DV}
}

@inproceedings{wang+2025visapp,
author={Yushan Wang and Shuhei Tarashima and Norio Tagawa},
title="{Efficient 3D Human Pose and Shape Estimation Using Group-Mix Attention in Transformer Models}",
booktitle={VISAPP},
year={2025},
}
}

\end{document}